%% file: main.tex
\documentclass[twoside]{article}

\usepackage[accepted]{aistats2024}
\usepackage{amsmath,amssymb,amsfonts}
\usepackage{algorithmic}
\usepackage{graphicx}
\usepackage{textcomp}
\usepackage[dvipsnames]{xcolor}

\usepackage{algorithmic}
\usepackage{amsmath}
\usepackage{graphicx}
\usepackage{color} 
\usepackage{algorithm} 
\usepackage{subcaption}
\usepackage{multirow}
\usepackage{multicol}
\usepackage[font=footnotesize]{caption}
\usepackage{tabularx,booktabs}
\usepackage{url}
\usepackage{hyperref}

\usepackage[round]{natbib}

\newcommand{\boldhdr}[1]{\noindent \textbf{#1.}}

\usepackage{comment}
\usepackage{ifthen}
\newcommand{\showcomments}{yes}
\newcommand\m[1]{
    \ifthenelse{\equal{\showcomments}{yes}}{{\color{red} [Ming: #1]}}{\ignorespaces}
}
\newcommand\ming[1]{
    \ifthenelse{\equal{\showcomments}{yes}}{{\color{red} [Ming: #1]}}{\ignorespaces}
}
\newcommand{\kai}[1]{\textcolor{black}{#1}} 

\begin{document}

%

%

\twocolumn[

\aistatstitle{Self-Supervised Quantization-Aware Knowledge Distillation}

\aistatsauthor{ Kaiqi Zhao \And Ming Zhao }

\aistatsaddress{ Arizona State University \And Arizona State University } ]

\input{abstract}
\input{introduction}
\input{background}

\input{methodology}
\input{evaluation}
\input{ablation}
\input{conclusion}
\input{acknowledgments}

\bibliographystyle{plainnat}
\bibliography{main}

\appendix
\renewcommand{\thesection}{\arabic{section}} 
\input{supplement}

\end{document}

%% file: abstract.tex
\begin{abstract}

Quantization-aware training (QAT) and Knowledge Distillation (KD) are combined to achieve competitive performance in creating low-bit deep learning models.
However, existing works applying KD to QAT require tedious hyper-parameter tuning to balance the weights of different loss terms, assume the availability of labeled training data, and require complex, computationally intensive training procedures for good performance.
To address these limitations, this paper proposes a novel Self-Supervised Quantization-Aware Knowledge Distillation (SQAKD) framework.
SQAKD first unifies the forward and backward dynamics of various quantization functions, making it flexible for incorporating various QAT works.
Then it formulates QAT as a co-optimization problem that simultaneously minimizes the KL-Loss between the full-precision and low-bit models for KD and the discretization error for quantization, without supervision from labels. 
A comprehensive evaluation shows that SQAKD substantially outperforms the state-of-the-art QAT and KD works for a variety of model architectures. 
\kai{Our code is at: \url{https://github.com/kaiqi123/SQAKD.git}.}

\end{abstract}


%% file: introduction.tex
\section{Introduction}

%

Deep neural networks (DNNs) have substantial computational and memory requirements. As the use of deep learning grows rapidly on a wide variety of Internet of Things and devices, the mismatch between resource-hungry DNNs and resource-constrained devices also becomes increasingly severe~\citep{zhang2023towards, zhao2023automatic}.
Quantization is one of the important model compression approaches to address this challenge by converting the full-precision model weights or activations to lower precision.
In particular, Quantization-Aware Training (QAT)~\citep{krishnamoorthi2018quantizing} has achieved promising results in creating low-bit models, which starts with a pre-trained model and performs quantization during retraining. 
%
However, most QAT works lead to considerable accuracy loss due to quantization, and no algorithm achieves consistent performance on every model architecture (e.g., VGG, ResNet, MobileNet)~\citep{li2021mqbench}.
Also, the various QAT works are motivated by different intuitions and lack a commonly agreed theory, which makes it challenging to generalize.
Moreover, we empirically find that they do not work well on low-bit networks (1-3 bits). 
Therefore, we argue that there is a great need for a generalized, simple yet effective framework that is flexible to incorporate and improve various QAT algorithms for both low-bit and high-bit quantization. 

Recent works apply Knowledge Distillation (KD) to QAT to mitigate the accuracy loss of the low-precision networks (termed ``student'') by transferring knowledge from full-precision or high-precision networks (termed ``teacher'') during training. 
However, these KD-applied QAT works 
1) require tedious hyper-parameter tuning to balance the weights of different loss terms;
2) assume the availability of labeled training data which is in practice difficult and sometimes even infeasible to obtain; 
3) require complex, computationally intensive training procedures for good performance (see Section~\ref{sec:related_works_KD_and_QAT} for details); 
and 4) focus narrowly on one specific KD approach alongside one particular quantizer which, as shown in our study, does not consistently perform well. 
In this work, we propose a simple yet effective framework --- Self-Supervised Quantization-Aware Knowledge Distillation (SQAKD). 
SQAKD first unifies the forward and backward dynamics of various quantization functions and shapes quantization-aware training as an optimization problem minimizing the discretization error between original weights/activations and their quantized counterparts.
Then, we perform an in-depth analysis of the loss landscape of QAT where the conventional training is replaced by KD.
Our analysis reveals that CE-Loss (i.e., cross-entropy loss with labels) does not cooperate effectively with KL-Loss (i.e., the Kullback-Leibler divergence loss between the teacher’s and student’s penultimate outputs) and their combination may degrade the network performance.
Thus, SQAKD proposes the novel formulation of QAT as a co-optimization problem that simultaneously minimizes the KL-Loss between the full-precision and low-bit models for KD and the discretization error for quantization without supervision from labels. 
%

Compared to existing QAT methods and those that combine KD and QAT, the proposed SQAKD has several advantages. 
First, SQAKD is flexible for incorporating various QAT works by unifying the optimization of their forward and backward dynamics.
Second, SQAKD improves the SOTA QAT works in both convergence speed and accuracy by using the full-precision teacher's help to guide gradient updates for low-bit weights, and it simultaneously minimizes the KL-Loss and the discretization error, making the estimated discrete gradients close to the continuous gradients. 
Third, SQAKD is hyperparameter-free, since it uses only the KL-Loss as the training loss and does not require the weight adjustment of multiple loss terms that appear in previous methods. 
Fourth, SQAKD operates in a self-supervised manner without labeled data, supporting a broad range of applications in practical scenarios.
Finally, SQAKD offers simplicity in training procedures by requiring only one training phase to update the student, which reduces training cost and promotes usability and reproducibility. 



Our comprehensive evaluation shows that SQAKD outperforms SOTA QAT and KD works significantly on various model architectures (VGG, ResNet, MobileNet-V2, ShuffleNet-V2, and SqueezeNet).
First, compared to QAT, e.g., EWGS, PACT, LSQ, and DoReFa, SQAKD improves their convergence speed and top-1 accuracy (by up to 15.86\%) for 1-8 bit quantization.
%
Second, compared to 11 KD methods, SQAKD outperforms them by up to 17.09\% on 1-bit VGG-13 with CIFAR-100.
%
Third, compared to KD-integrated QAT, SQAKD achieves the smallest accuracy drop, outperforming the baselines by up to 3.06\% on 2-bit ResNet-32 with CIFAR-100.
Furthermore, on Jetson Nano hardware, SQAKD achieves an inference speedup of 3$\times$ for 8-bit quantization on Tiny-ImageNet.




In summary, our main contributions are summarized as: First, we are the first to 1) quantitatively investigate and benchmark 11 KD methods in the context of QAT, and 2) provide an in-depth analysis of the loss landscape of KD within QAT.
Second, we propose a Self-Supervised Quantization-Aware Knowledge Distillation (SQAKD) method which operates in a self-supervised manner without labeled data and eliminates the need for hyper-parameter balancing. 
It is applicable to incorporate various quantizers, and it consistently outperforms state-of-the-art QAT, KD, and KD-integrated QAT methods on various models and datasets.
Third, we open source all quantized networks, including those without any accuracy loss, such as 2-bit VGG-8, 4-bit ResNet-32, and 8-bit MobileNet-V2 that get 91.55\%, 71.65\%, and 58.13\% in top-1 accuracy on CIFAR-10, CIFAR-100, and Tiny-ImageNet, respectively. These low-precision networks are beneficial for diverse real-world applications.

%% file: background.tex
\section{Background and Related Works}


\begin{table*}[t]
\scriptsize
\centering
\caption{
\kai{Summary of related works that apply KD to QAT.
The training cost is estimated by the sum of the initial training time for the pre-trained teachers ($N \times T_{pre}$), where $N$ is the number of pre-trained teachers and $T_{pre}$ is each teacher's training cost, and the training time during QAT with KD for both the teacher ($M_t \times T_t$) and the student ($M_s \times T_s$). Here, $M_t$ and $M_s$ denote the number of training phases for the teacher and student, respectively, while $T_t$ and $T_s$ represent their respective training costs per phase.}
}
\label{tab:summary_related_works}
\vspace{-10pt}
\setlength{\tabcolsep}{2.5pt}
\begin{tabularx}{0.99\textwidth}{@{}lllllll@{}}
\toprule
               & \begin{tabular}[c]{@{}l@{}}Self-\\ Supervised\end{tabular} & \begin{tabular}[c]{@{}l@{}}Number\\ of hyper-\\ parameters\\ for balancing \\ loss terms\end{tabular} & \begin{tabular}[c]{@{}l@{}}Number of \\ training \\ phases\end{tabular} & Training cost            & Initial teacher type                  & \begin{tabular}[c]{@{}l@{}}Train jointly / \\ Train the student only\end{tabular} \\ \midrule
AP-SCHEME-A    & No                                                         & 3                                                                                                     & 1                                                                       & $T_{s} + T_{t}$          & One random initialized full-precision & Train jointly                                                                     \\
AP-SCHEME-B    & No                                                         & 3                                                                                                     & 1                                                                       & $T_{pre}+T_{s}$          & One pre-trained full-precision        & Train the student only                                                            \\
AP-SCHEME-C    & No                                                         & 3                                                                                                     & 2                                                                       & $T_{pre} + 2T_{s}$      & One pre-trained full-precision        & Train the student only                                                            \\
QKD            & No                                                         & 2                                                                                                     & 3                                                                       & $T_{pre}+3T_{s}+T_{t}$  & One pre-trained full-precision        & Train jointly                                                                     \\
CMT-KD         & No                                                         & 3                                                                                                     & 1                                                                       & $NT_{pre}+T_{s}+T_{t}$  & Multiple different bit-widths         & Train jointly                                                                     \\
SPEQ           & No                                                         & 1                                                                                                     & 1                                                                       & $T_{pre}+T_{s}+T_{t}$    & One quantized with target bit-width   & Train jointly                                                                     \\
PTG   & No                                                         & 2                                                                                                     & 4                                                                       & $T_{pre}+4T_{s}+T_{t}$  & One pre-trained full-precision        & Train jointly                                                                     \\
\textbf{SQAKD (Ours)} & \textbf{Yes}                                               & \textbf{0}                                                                                            & \textbf{1}                                                              & $\mathbf{T_{pre}+T_{s}}$ & One pre-trained full-precision        & Train the student only                                                            \\ \bottomrule
\end{tabularx}
\end{table*}



\smallskip
\boldhdr{Quantization}
There are two main approaches to quantization.
Post-Training Quantization (PTQ)~\citep{li2021brecq} quantizes a pre-trained model without retraining and often leads to worse accuracy degradation than QAT~\citep{krishnamoorthi2018quantizing}, which performs quantization during retraining. 
This paper focuses on QAT.
Many QAT works emphasize designing the forward and backward propagation of the quantizer, the function that converts continuous weights or activations to discrete values.
Early works such as BNN~\citep{courbariaux2016binarized} and XNOR-Net~\citep{rastegari2016xnor} employ channel-wise scaling in the forward pass, while DoReFa-Net~\citep{zhou2016dorefa} introduces a universal scalar for all filters.
Recent works utilize trainable parameters for the quantizer, improving control over facets such as clipping ranges (e.g., PACT~\citep{choi2018pact}, LSQ~\citep{esser2019learned}, APoT~\citep{li2019additive}, and DSQ~\citep{gong2019differentiable}), i.e., the bounds where the input values are constrained, and quantization intervals (e.g., QIL~\citep{jung2019learning} and EWGS~\citep{lee2021network}), i.e., the step size between adjacent quantization levels.
However, existing QAT methods lead to different levels of accuracy loss and are motivated by various heuristics, lacking a commonly agreed theory.
Furthermore, MQbench~\citep{li2021mqbench} reveals that the difference between QAT algorithms is not as substantial as reported in their original papers;
and no algorithm achieves the best performance on all architectures. 
%
Moreover, most QAT algorithms focus on high-bit networks (4 bits or more), underperforming on low-bit networks (see Section~\ref{sec:improvement_on_quantization} for details). 
Therefore, this paper aims to close this gap, providing a generalized framework for integrating and improving various QAT works across both low-bit and high-bit precisions.
\smallskip
\boldhdr{Knowledge Distillation (KD)}
KD transfers knowledge from large networks (termed ``teacher'') to improve the performance of small networks (termed ``student'').
%
%
Hinton et al. first proposed to transfer soft logits by minimizing the KL divergence between the teacher's and student's softmax outputs and the cross-entropy loss with data labels~\citep{hinton2015distilling}. 
%
Later, some works~\citep{tung2019similarity,peng2019correlation,zhao2023contrastive} proposed to transfer various forms of intermediate representations, such as FSP matrix~\citep{yim2017gift} and attention maps~\citep{zagoruyko2016paying}.
%
However, despite the success of KD for image classification, the understanding of KD in model quantization is limited. 







\smallskip
\boldhdr{Knowledge Distillation + Quantization}
\label{sec:related_works_KD_and_QAT}
Several recent works employ KD to mitigate accuracy loss from quantization, with the low-precision network as the student and the high- or full-precision network as the teacher.
Mishra et al. proposed three schemes in Apprentice (AP) to improve the performance of ternary-precision or 4-bit networks.
QKD~\citep{kim2019qkd} coordinates quantization and KD through three phases, including self-studying, co-studying, and tutoring.
SPEQ~\citep{boo2021stochastic} constructs a teacher using the student's parameters and applies stochastic bit precision to the teacher's activations.
PTG~\citep{zhuang2018towards} proposed a four-stage training strategy, emphasizing sequential optimization of quantized weights and activations, progressive reduction of bit width, and concurrent training of the teacher and the student.
CMT-KD~\citep{pham2023collaborative} encourages collaborative learning among multiple quantized teachers and mutual learning between teachers and the student.

We summarize the aforementioned works in Table~\ref{tab:summary_related_works}. 
Despite their contributions, they have several limitations.
First, they all require a delicate balance for the weights of different loss terms, which leads to time-consuming and error-prone hyperparameter tuning. 
For example, QKD requires three hyperparameters to balance the loss terms and their influence on performance is substantial.
%
Second, they all assume that labels of the training data are always available during training, whereas in real-world scenarios access to labeled data is often limited and/or costly.
%
Third, they require a complex, computationally expensive training process; for example, 3 training phases and 4 optimization strategies are required in QKD and PTG, respectively, costing training time and usability.
%
Finally, these works focus on one specific KD approach alongside one particular quantizer, which cannot work well consistently for different applications on different hardware~\citep{li2021mqbench}. 
%
This paper proposes a novel quantization-aware KD framework that eliminates hyperparameter balancing, operates in a self-supervised manner, supports diverse quantizers, and offers simplicity in training procedures with reduced training costs, as discussed in the rest of the paper.



%% file: methodology.tex
\section{Methodology}

\begin{figure}[t]
    \centering
    \includegraphics[width=\linewidth]{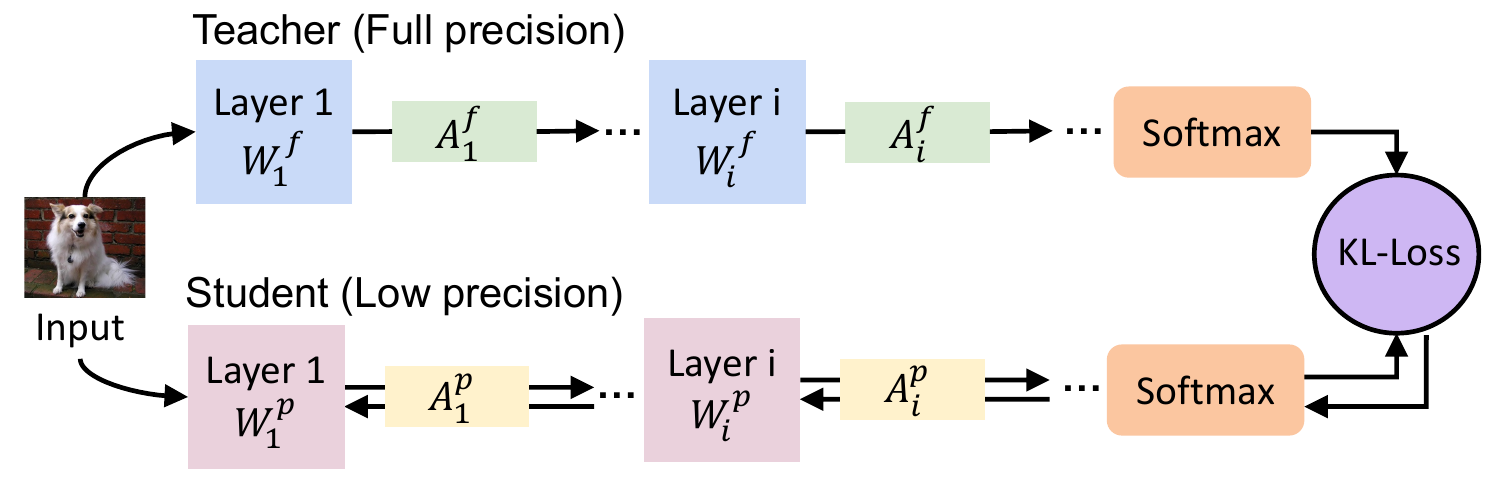}
    \vspace{-20pt}
    \caption{\scriptsize{Workflow of SQAKD.}}
    \label{fig:workflow}
\end{figure}





Figure~\ref{fig:workflow} illustrates the workflow of the proposed framework --- Self-Supervised Quantization-Aware Knowledge Distillation (SQAKD). 

\subsection{QAT as Constrained Optimization}
To provide a generalized theoretic framework, SQAKD first unifies the forward and backward dynamics of various quantizers and formulates QAT as an optimization problem.

\smallskip
\boldhdr{Forward Propagation}
\kai{Let us define $Quant(\cdot)$ as a uniform quantizer that converts a full-precision input $x$ to a quantized output $x_q = Quant(x)$.
$x$ can be the activations or weights of the network.
First, the quantizer $Quant(\cdot)$ applies a clipping function $Clip(\cdot)$, which normalizes and restricts the full-precision input $x$ to a limited range, producing a full-precision latent presentation $x_c$, as follows:}
\vspace{-6pt}
\begin{equation}\label{equ:clip_function}
\small
    x_c = Clip(x, \{p_i\}_{i=1}^{i=K_c}, v, m),
\vspace{-6pt}
\end{equation}
where $v$ and $m$ are the lower and upper bounds of the range, respectively,
$\{p_i\}_{i=1}^{i=K_c}$ denotes the set of trainable parameters needed for quantization,
and $K_c$ denotes the number of parameters.
%
Note that different quantizers have different schemes for $Clip(\cdot)$.
For example, in PACT, the lower bound $v$ is set to 0 and the upper bound $m$ is a trainable parameter optimized during training. 
The quantizer requires only one parameter, that is, $\{p_1|p_1=m, K_c=1\}$,
and the clipping function is described as: $x_c = Clip(x, \{p_1|p_1=m\}, 0, m) = 0.5(|x|-|x-m|+m)$.
%
In EWGS, $v$ and $m$ are set to 0 and 1, respectively, and every quantized layer uses separate parameters (i.e. $p_1$ and $p_2$) for quantization intervals: $x_c = Clip(x, \{p_1, p_2\}, 0, 1) = clip(\frac{x-p_1}{p_1-p_2}, 0, 1)$.

Then, the quantizer $Quant(\cdot)$ converts the clipped value $x_c$ to a discrete quantization point $x_q$ using the function $R(\cdot)$ that contains a round function:
\vspace{-5pt}
\begin{equation}\label{equ:round_function}
\small
    x_q = R(x_c, b, \{q_i\}_{i=1}^{i=K_r}),
\vspace{-5pt}
\end{equation}
where $b$ is the bit width and $\{q_i\}_{i=1}^{i=K_r}$ denotes the set of trainable parameters.
Note that $\{q_i\}_{i=1}^{i=K_r}$ is not necessary for some quantizers. 
For example, in EWGS, if activations are the input, $x_q = R(x_c, b) = \frac{round((2^b-1) \cdot x_c)}{2^b-1}$; and if weights are the input, $x_q = R(x_c, b) = 2(\frac{round((2^b-1) \cdot x_c)}{2^b-1} - 0.5)$.
%
In some quantizers, like PACT and LSQ, the trainable parameters in the function $R(\cdot)$ are the same as those in the clipping function $Clip(\cdot)$, that is, $\{q_i|q_i=p_i\}_{i=1}^{i=K_r}$ and $K_r=K_c$.
%

In summary, the quantizer $Quant(\cdot)$ is described as: $x_q = Quant(x, \alpha, b, v, m)$,
where $\alpha$ denotes a shorthand for the set of all the parameters in the functions $R(\cdot)$ and $Clip(\cdot)$: $\alpha = \{ \{p_i\}_{i=1}^{i=K_c}, \{q_i\}_{i=1}^{i=K_r} \}$.

\smallskip
\boldhdr{Backward Propagation}
Directly training a quantized network using back-propagation is impossible since the quantizer $Q(\cdot)$ is non-differentiable.
This issue arises due to the round function in Eq.~\ref{equ:round_function}, which produces near-zero derivatives almost everywhere.
To solve this problem, most QAT works use Straight-Through Estimator (STE)~\citep{bengio2013estimating} to approximate the gradients: $\frac{\partial L}{\partial x_c} = \frac{\partial L}{\partial x_q}$.
Instead of the commonly used STE for backpropagation, we propose a novel formula to integrate the discretization error ($x_c - x_q$), representing the deviation between full-precision and its quantized weights/activations:
%
\vspace{-6pt}
\begin{equation}\label{equ:back_propogation}
\small
    \frac{\partial L}{\partial x_c} = \frac{\partial L}{\partial x_q} + \mu \cdot (x_c - x_q),
\vspace{-6pt}
\end{equation}
with $\mu$ being a non-negative value.
STE is represented by setting $\mu$ to zero, while EWGS is obtained when $\mu$ is the product of $\delta$ (a non-negative value), $sign(\frac{\partial L}{\partial x_q})$, and $\frac{\partial L}{\partial x_q}$.
Note that $\mu$ can also be updated by other schemes, like Curriculum Learning driven strategy~\citep{bengio2009curriculum,zhao2022knowledge}.

\smallskip
\boldhdr{Optimization Objective}
We frame QAT as an optimization problem minimizing both the discretization error as well as the discrepancy between model predictions and true labels.
The goal is precise quantization without sacrificing predictive accuracy. 
Thus, the optimization objection is defined as follows:
\vspace{-3pt}
\begin{equation}\label{equ:optimization}
    \small
    \begin{split}
    & \quad \quad \quad \underset {W_f, \alpha_W, \alpha_A} {min} L(W_f)\\
    s.t. \quad  &W_q = Quant_W(W_f, \alpha_W, b_W, v_W, m_W) \\
    &A_q = Quant_A(A_f, \alpha_A, b_A, v_A, m_A),  
    \end{split}
\vspace{-6pt}
\end{equation}
where $Quant_W(\cdot)$ and $Quant_A(\cdot)$ are the quantizers for weights and activations,
and $W_f$/$A_f$ and $W_q$/$A_q$ are the model's full-precision and quantized weights/activations.
Note that $L(\cdot)$ can be the cross entropy loss with labels or other loss functions, e.g., distillation loss (see Sections~\ref{sec:analysis_of_kd_on_quantization} and ~\ref{sec:SQAKD} for details). 

To the best of our knowledge, we are the first to generalize diverse quantizers, encompassing forward and backward propagations into a unified formulation of an optimization problem.
This generalization and formulation enable our framework to be flexible in integrating various SOTA quantizers and effectively improve their performance (see Section~\ref{sec:improvement_on_quantization} for the results).


\subsection{Analysis of KD in QAT} 
\label{sec:analysis_of_kd_on_quantization}





\smallskip
\noindent \textbf{Does KD perform well in QAT?}
%
Despite the wide use of KD, its effectiveness in addressing the QAT problem lacks a thorough study.
To apply KD in QAT, we let a pre-trained full-precision network act as the teacher, guiding a low-bit student with the same architecture.
Then, the training loss in Eq.~\ref{equ:optimization} is defined as a linear combination of the cross-entropy loss with labels and the distillation loss between the teacher's and student's output distributions, controlled by a hyper-parameter $\lambda$: 
\vspace{-3pt}
\begin{equation}\label{equ:optimization_ce_distill}
\small
    L = (1-\lambda) L_{CE} + \lambda L_{Distill}.
\vspace{-3pt}
\end{equation}
Note that the distillation loss $L_{Distill}$ can be one term, e.g., the KL divergence loss~\citep{hinton2015distilling}, or multiple terms, e.g., the intermediate-presentation-based contrastive losses~\citep{zhao2023contrastive}.
We hypothesize that existing KD methods, while effective in normal training scenarios, may not achieve satisfactory performance in QAT.
This is because quantized networks exhibit lower representational capacity compared to their full-precision counterparts~\citep{nahshan2021loss}, making it difficult to optimize multiple loss terms effectively.
In addition, quantization introduces additional noise to the network weights or activations due to discretization. 
This noise can degrade the performance of the KD methods that rely on fine-grained information or target matching specific outputs.
%
To validate this hypothesis, we comprehensively evaluate 11 KD methods in the context of quantization (see Section~\ref{sec:comparison_with_KD} for evaluation results).
Notably, this evaluation represents the first attempt, to the best of our knowledge, to provide a thorough assessment of the performance of KD in addressing QAT problems.

\noindent \textbf{Are both the cross-entropy loss and distillation loss necessary?}
%
To address the issue of KD methods underperforming in QAT, we analyze the significance of two loss components, shown in Eq.~\ref{equ:optimization_ce_distill}: the cross-entropy loss (termed CE-Loss) with labels and the KL divergence loss (termed KL-Loss) between the teacher's and student's penultimate outputs (before logits).
We focus specifically on KL-Loss rather than other forms of distillation loss, as the conventional KD~\citep{hinton2015distilling}, using KL-Loss, performs the best among existing KD works~\citep{tian2019contrastive}.
CE-Loss quantifies the alignment between the low-bit network's predictions and the true labels, while KL-Loss measures the similarity between the low-bit network and the pre-trained full-precision network. 

%

\begin{figure}[t]
  \centering
  \begin{subfigure}{.45\linewidth}
    \centering
      \includegraphics[width=4.cm]{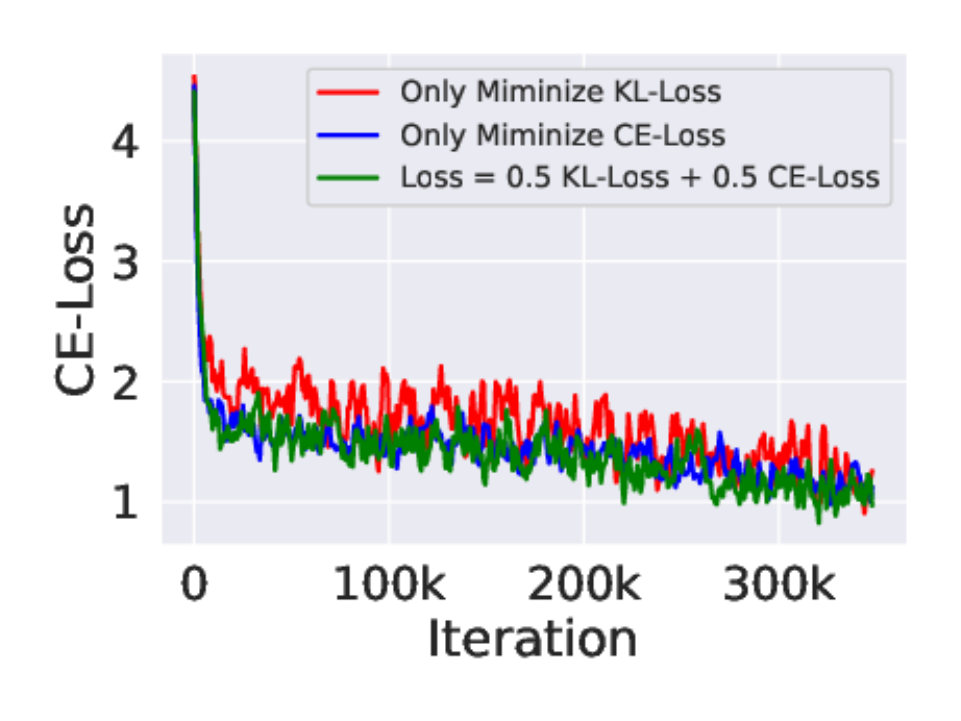}
    \vspace{-20pt}
    \caption{CE-Loss}
    \label{fig:loss_analysis_ce}
  \end{subfigure}
  \begin{subfigure}{.45\linewidth}
    \centering
    \includegraphics[width=4.cm]{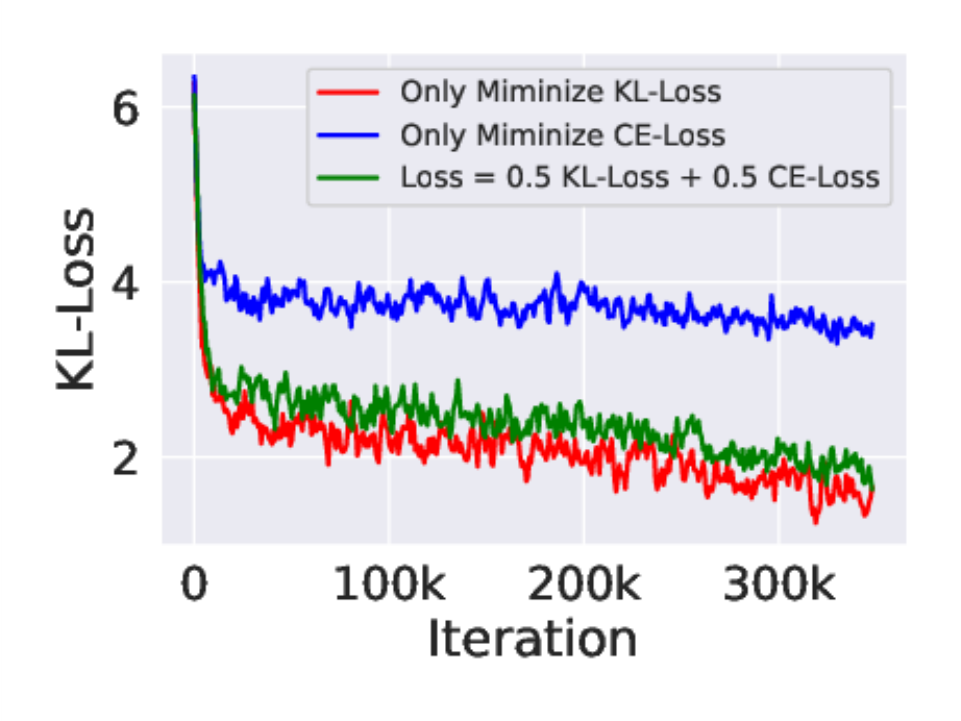}
    \vspace{-20pt}
    \caption{KL-Loss}
    \label{fig:loss_analysis_kl}
  \end{subfigure}
   \vspace{-7pt}
  \caption{Illustration of the evolution of (a) CE-Loss and (b) KL-Loss in each iteration during the training of 1-bit VGG-13 on CIFAR-100.}
   \label{fig:loss_analysis}
\end{figure}

We consider three scenarios: 1) only minimizing KL-Loss by setting $\lambda=1$ in Eq.~\ref{equ:optimization_ce_distill}, 2) only minimizing CE-Loss ($\lambda=0$), 
and 3) jointly minimizing CE-Loss and KL-Loss with equal weights ($\lambda=0.5$).
Figure~\ref{fig:loss_analysis} illustrates the evolution of CE-Loss and KL-Loss in each iteration during the training of 1-bit VGG-13 on CIFAR-100. 
We observe Figure~\ref{fig:loss_analysis_ce} that solely minimizing KL-Loss effectively leads to the concurrent minimization of both CE-Loss and KL-Loss.
This indicates that the low-bit student can produce predictions close to the ground truth even without label supervision.
In comparison, as shown in Figure~\ref{fig:loss_analysis_kl}, solely minimizing CE-Loss or jointly minimizing CE-Loss and KL-Loss cannot achieves good minimization of KL-Loss.
This indicates that integrating CE-Loss into the loss function, either partially ($\alpha=0.5$) or entirely ($\alpha=0$), negatively impacts the performance.
Therefore, we propose that solely minimizing KL-Loss is sufficient to achieve optimal gradient updates in the quantized network;
CE-Loss does not cooperate effectively with KL-Loss, and their combination may actually degrade the network performance.
Also, by excluding CE-Loss and considering only KL-Loss, the loss function becomes hyperparameter-free and thus eliminates the need for balancing the loss terms. 


\subsection{Optimization via Self-Supervised KD}
\label{sec:SQAKD}

As we drop the CE-Loss and keep only the KL-Loss in SQAKD,
the optimization objection in Eq.~\ref{equ:optimization} becomes:
\vspace{-6pt}
\begin{equation}\label{equ:optimization_SQAKD}
    \small
    \begin{split}
    \underset {W^S_f, \alpha_W, \alpha_A} {min} 
    &KL (S(h^T/\rho) || S(h^S/\rho)) \\
    s.t. \quad &W_q^S = Quant_W(W^S_f, \alpha_W, b_W, v_W, m_W) \\
    &A_q^S = Quant_A(A^S_f, \alpha_A, b_A, v_A, m_A),
    \end{split} 
\vspace{-6pt}
\end{equation}
where $\rho$ is the temperature, which makes distribution softer for using the dark knowledge,
$Y$ is the ground-truth labels, and 
$h^T$ and $h^S$ are the penultimate layer outputs of the teacher and student, respectively.
$Quant_W(\cdot)$ and $Quant_A(\cdot)$ are the quantization function for the student's weights and activations,
and $W^S_f$/$A^S_f$ and $W^S_q$/$A^S_q$ are the student's full-precision weights/activities and quantized weights/activities.

During training, the teacher's weights are frozen and it performs the forward propagation only.
In the forward pass, the student's parameters are quantized, while the corresponding full-precision values are preserved internally. 
During the backward propagation, the gradients are applied to the student's preserved full-precision values.
%
After convergence, the student retains its full-precision weights and the parameters utilized in the quantizer.
The student's quantized weights are obtained by applying the quantizer to the full-precision weights.
\kai{The overall procedure of SQAKD is summarized in Algorithm~\ref{alg:SQAKD}.}

\begin{algorithm}[t]
    \centering
    \small
    \caption{\kai{Self-Supervised Quantization-Aware Knowledge Distillation}}
    \begin{algorithmic}[1]
        \STATE \textbf{Input:} A pre-trained full-precision model $F_f^T$, and target bit-widths $b_W$ for weights and $b_A$ for activations.
        \STATE \textbf{Output:} A quantized model $F_q^S$ with specified bit-widths $b_W$ for weights and $b_A$ for activations.
        \STATE \textbf{Parameters:} Lower and upper bounds $v_W, m_W$ for weights, and $v_A, m_A$ for activations; quantization function parameters $\alpha_W = \{ \{p_{Wi}\}_{i=1}^{i=K_c}, \{q_{Wi}\}_{i=1}^{i=K_r} \}$ for weights and $\alpha_A = \{ \{p_{Ai}\}_{i=1}^{i=K_c}, \{q_{Ai}\}_{i=1}^{i=K_r} \}$ for activations.
        \STATE \textbf{Hyperparameters:} Temperature $\rho$
        \STATE \textbf{Initialization:} Initialize the student with the pre-trained teacher: $F_f^S = F_f^T$
        \FOR {each training iteration given every input $X$}
            \STATE \textbf{Forward Propagation:} 
               \STATE Student: $h^S = F_f^S(X)$
               \STATE $\begin{aligned} 
               s.t. \quad &W_c = Clip_W(W^S_f, \{p_{Wi}\}_{i=1}^{i=K_c}, v_W, m_W) \quad (Eq.~\ref{equ:clip_function}) \\
               &W_q^S = R_W(W_c, b_W, \{q_{Wi}\}_{i=1}^{i=K_r}) \quad (Eq.~\ref{equ:round_function})\\
               \quad &A_c = Clip_A(A^S_f, \{p_{Ai}\}_{i=1}^{i=K_c}, v_A, m_A) \quad (Eq.~\ref{equ:clip_function})\\
               &A_q^S = R_A(A_c, b_A, \{q_{Ai}\}_{i=1}^{i=K_r}) \quad (Eq.~\ref{equ:round_function})
               \end{aligned}$
                \STATE Teacher: $h^T = F_f^T(X)$
               \STATE \textbf{Minimize the loss function:}
                \STATE $ L = KL (S(h^T/\rho) || S(h^S/\rho))$ (Eq.~\ref{equ:optimization_SQAKD})
               \STATE \textbf{Backward Propagation:} 
               \STATE Calculate the gradient of discrete weights $W_q$ via backpropagation: $\frac{\partial L}{\partial W_q}$.
              \STATE Calculate the gradient of latent values $W_c$: $\frac{\partial L}{\partial W_c} = \frac{\partial L}{\partial W_q} + \mu \cdot (W_c - W_q)$ (Eq.~\ref{equ:back_propogation}).
              \STATE Propagate the gradient to the input.
        \ENDFOR
    \end{algorithmic}
\label{alg:SQAKD}
\end{algorithm}


%% file: evaluation.tex
\section{Evaluation}


\begin{table}[t]
\scriptsize
\centering
\caption{Models and datasets.}
\vspace{-10pt}
\label{tab:models_datasets}
\setlength{\tabcolsep}{1.0pt}
\begin{tabular}{l|p{7.2cm}}
\toprule
Model   & \textbf{ResNet}~\citep{he2016identity}, \textbf{VGG}~\citep{eigen2015predicting}, \textbf{MobileNet}~\citep{sandler2018mobilenetv2}, \textbf{ShuffleNet}~\citep{zhang2018shufflenet},  \textbf{SqueezeNet}~\citep{iandola2016squeezenet}, \textbf{AlexNet}~\citep{krizhevsky2012imagenet} \\ \midrule
Dataset & \textbf{CIFAR-10}~\citep{krizhevsky2009learning}, \textbf{CIFAR-100}~\citep{krizhevsky2009learning}, \textbf{Tiny-ImageNet}~\citep{le2015tiny} \\ 
\bottomrule
\end{tabular}
\end{table}


\begin{table}[t]
\scriptsize
\centering
\caption{Baselines.}
\vspace{-10pt}
\label{tab:baselines}
\setlength{\tabcolsep}{2.5pt}
\begin{tabularx}{0.48\textwidth}{@{}l|l@{}}
\toprule
Group       & Baselines \\ \midrule
QAT            & \begin{tabular}[c]{@{}l@{}}\textbf{PACT}~\citep{choi2018pact}, \textbf{LSQ}~\citep{esser2019learned}, \\ \textbf{DoReFa}~\citep{zhou2016dorefa}, \textbf{EWGS}~\citep{lee2021network}\end{tabular}          \\ \midrule

KD             & \begin{tabular}[c]{@{}l@{}}\textbf{SP}~\citep{tung2019similarity}, \textbf{AT}~\citep{zagoruyko2016paying}, \\
\textbf{FitNet}~\citep{romero2014fitnets}, \textbf{CC}~\citep{peng2019correlation}, \\
\textbf{VID}~\citep{ahn2019variational}, \textbf{RKD}~\citep{park2019relational}, \\
\textbf{AB}~\citep{heo2019knowledge}, \textbf{FT}~\citep{kim2018paraphrasing}, \\
\textbf{FSP}~\citep{yim2017gift}, \textbf{NST}~\citep{huang2017like}, \\\textbf{CRD}~\citep{tian2019contrastive}, \textbf{CKTF}~\citep{zhao2023contrastive}\end{tabular}          \\ \midrule

\begin{tabular}[c]{@{}l@{}}KD+QAT\end{tabular} & \begin{tabular}[c]{@{}l@{}} \textbf{SPEQ}~\citep{boo2021stochastic}, \textbf{PTG}~\citep{zhuang2018towards} \\ \textbf{QKD}~\citep{kim2019qkd}, \textbf{CMT-KD}~\citep{pham2023collaborative} \end{tabular}          \\ \bottomrule
\end{tabularx}
\end{table}

We conducted an extensive evaluation on diverse models and datasets, as listed in Table~\ref{tab:models_datasets} (see Appendix for more details).
We compared SQAKD with three groups of state-of-the-art (SOTA) methods: standalone QAT, KD, and KD-integrated QAT, as listed in Table~\ref{tab:baselines}.

We implemented SQAKD on PyTorch version 1.10.0 and Python version 3.9.7.
We used four Nvidia RTX 2080 GPUs for model training and conduct inference experiments on Jetson Nano using NVIDIA TensorRT.
Please see Appendix for more implementation details.

\subsection{Improvements on SOTA QAT Methods}
\label{sec:improvement_on_quantization}

\begin{table*}[t]
\scriptsize
\centering
\caption{Top-1 test accuracy (\%) on CIFAR-10 and CIFAR-100. ``FP'' denotes the full-precision model's accuracy. ``W$*$A$\times$'' denotes that the weights and activations are quantized into $*$bit and $\times$bit, respectively. The number inside the parenthesis is the improvement compared to EWGS.}
\vspace{-10pt}
\label{tab:cifar_improvement_on_quantization}
\setlength{\tabcolsep}{2.0pt}
\begin{tabularx}{0.99\textwidth}{@{}l|llllll|llllll@{}}
\toprule
Dataset               & \multicolumn{6}{c|}{CIFAR-10}                                                                                                                                                                                                                                                                                                                                                                                                        & \multicolumn{6}{c}{CIFAR-100}                                                                                                                                                                                                                                                                                                                                                                                                        \\ \midrule
Model                 & \multicolumn{3}{c|}{\begin{tabular}[c]{@{}c@{}}VGG-8\\ (FP: 91.27)\end{tabular}}                                                                                                                                            & \multicolumn{3}{c|}{\begin{tabular}[c]{@{}c@{}}ResNet-20\\ (FP: 92.58)\end{tabular}}                                                                                                                   & \multicolumn{3}{c|}{\begin{tabular}[c]{@{}c@{}}VGG-13\\ (FP: 76.36)\end{tabular}}                                                                                                                                           & \multicolumn{3}{c}{\begin{tabular}[c]{@{}c@{}}ResNet-32\\ (FP: 71.33)\end{tabular}}                                                                                                                    \\ \midrule
Bit-width                   & W1A1                                                             & W2A2                                                             & \multicolumn{1}{l|}{W4A4}                                                             & W1A1                                                             & W2A2                                                             & W4A4                                                             & W1A1                                                             & W2A2                                                             & \multicolumn{1}{l|}{W4A4}                                                             & W1A1                                                             & W2A2                                                             & W4A4                                                             \\ 
EWGS                  & 87.77                                                            & 90.84                                                            & \multicolumn{1}{l|}{90.95}                                                            & 86.42                                                            & 91.41                                                            & 92.40                                                             & 65.55                                                            & 73.31                                                            & \multicolumn{1}{l|}{73.41}                                                            & 59.25                                                            & 69.37                                                            & 70.50                                                             \\
\textbf{SQAKD (EWGS)} & \textbf{\begin{tabular}[c]{@{}l@{}}89.05\\ \textcolor{blue}{(+1.28)}\end{tabular}} & \textbf{\begin{tabular}[c]{@{}l@{}}91.55\\ \textcolor{blue}{(+0.71)}\end{tabular}} & \multicolumn{1}{l|}{\textbf{\begin{tabular}[c]{@{}l@{}}91.31\\ \textcolor{blue}{(+0.36)}\end{tabular}}} & \textbf{\begin{tabular}[c]{@{}l@{}}86.47\\ \textcolor{blue}{(+0.05)}\end{tabular}} & \textbf{\begin{tabular}[c]{@{}l@{}}91.80 \\ \textcolor{blue}{(+0.39)}\end{tabular}} & \textbf{\begin{tabular}[c]{@{}l@{}}92.59\\ \textcolor{blue}{(+0.19)}\end{tabular}} & \textbf{\begin{tabular}[c]{@{}l@{}}68.56\\ \textcolor{blue}{(+3.01)}\end{tabular}} & \textbf{\begin{tabular}[c]{@{}l@{}}74.65\\ \textcolor{blue}{(+1.34)}\end{tabular}} & \multicolumn{1}{l|}{\textbf{\begin{tabular}[c]{@{}l@{}}74.67\\ \textcolor{blue}{(+1.26)}\end{tabular}}} & \textbf{\begin{tabular}[c]{@{}l@{}}59.41\\ \textcolor{blue}{(+0.16)}\end{tabular}} & \textbf{\begin{tabular}[c]{@{}l@{}}69.99\\ \textcolor{blue}{(+0.62)}\end{tabular}} & \textbf{\begin{tabular}[c]{@{}l@{}}71.65\\ \textcolor{blue}{(+1.15)}\end{tabular}} \\ 
\bottomrule
\end{tabularx}
\end{table*}

\begin{table}[t]
\scriptsize
\centering
\caption{Top-1 test accuracy (\%) of ResNet-18 and VGG-11 on Tiny-ImageNet.}
\vspace{-10pt}
\label{tab:tiny_imagenet_improvement_on_quantization_large_models}
\setlength{\tabcolsep}{1.pt}
\begin{tabularx}{0.48\textwidth}{@{}l|lll|lll@{}}
\toprule
Model                   & \multicolumn{3}{c}{\begin{tabular}[c]{@{}c@{}}ResNet-18\\ (FP: 65.59)\end{tabular}}                                                                                                                   & \multicolumn{3}{c}{\begin{tabular}[c]{@{}c@{}}VGG-11\\ (FP: 59.47)\end{tabular}}                                                                                                                       \\ \midrule
Bit-width               & W3A3                                                             & W4A4                                                             & W8A8                                                             & W3A3                                                             & W4A4                                                             & W8A8                                                             \\
PACT                    & 58.09                                                            & 61.06                                                            & 64.91                                                            & 52.94                                                            & 57.10                                                            & 58.08                                                            \\
\textbf{\begin{tabular}[c]{@{}l@{}}SQAKD \\ (PACT)\end{tabular}}  & \textbf{\begin{tabular}[c]{@{}l@{}}61.34\\ \textcolor{blue}{(+3.25)}\end{tabular}} & \textbf{\begin{tabular}[c]{@{}l@{}}61.47\\ \textcolor{blue}{(+0.41)}\end{tabular}} & \textbf{\begin{tabular}[c]{@{}l@{}}65.78\\ \textcolor{blue}{(+0.87)}\end{tabular}} & \textbf{\begin{tabular}[c]{@{}l@{}}57.25\\ \textcolor{blue}{(+4.31)}\end{tabular}} & \textbf{\begin{tabular}[c]{@{}l@{}}59.05\\ \textcolor{blue}{(+1.95)}\end{tabular}} & \textbf{\begin{tabular}[c]{@{}l@{}}59.44\\ \textcolor{blue}{(+1.36)}\end{tabular}} \\ \midrule
LSQ                     & 61.99                                                            & 64.10                                                            & 65.08                                                            & 58.39                                                            & 59.14                                                            & 59.25                                                            \\
\textbf{\begin{tabular}[c]{@{}l@{}}SQAKD \\ (LSQ)\end{tabular}}    & \textbf{\begin{tabular}[c]{@{}l@{}}65.21\\ \textcolor{blue}{(+3.22)}\end{tabular}} & \textbf{\begin{tabular}[c]{@{}l@{}}65.34\\ \textcolor{blue}{(+1.24)}\end{tabular}} & \textbf{\begin{tabular}[c]{@{}l@{}}65.96\\ \textcolor{blue}{(+0.88)}\end{tabular}} & \textbf{\begin{tabular}[c]{@{}l@{}}58.43\\ \textcolor{blue}{(+0.04)}\end{tabular}} & \textbf{\begin{tabular}[c]{@{}l@{}}59.19\\ \textcolor{blue}{(+0.05)}\end{tabular}} & \textbf{\begin{tabular}[c]{@{}l@{}}59.42\\ \textcolor{blue}{(+0.17)}\end{tabular}} \\ \midrule
DoReFa                  & 61.94                                                            & 62.72                                                            & 63.23                                                            & 56.72                                                            & 57.28                                                            & 57.54                                                            \\
\textbf{\begin{tabular}[c]{@{}l@{}}SQAKD \\ (DoReFa)\end{tabular}} & \textbf{\begin{tabular}[c]{@{}l@{}}64.10\\ \textcolor{blue}{(+2.16)}\end{tabular}} & \textbf{\begin{tabular}[c]{@{}l@{}}64.56\\ \textcolor{blue}{(+1.84)}\end{tabular}} & \textbf{\begin{tabular}[c]{@{}l@{}}64.88\\ \textcolor{blue}{(+1.65)}\end{tabular}} & \textbf{\begin{tabular}[c]{@{}l@{}}57.02\\ \textcolor{blue}{(+0.3)}\end{tabular}}  & \textbf{\begin{tabular}[c]{@{}l@{}}58.93\\ \textcolor{blue}{(+1.65)}\end{tabular}} & \textbf{\begin{tabular}[c]{@{}l@{}}58.91\\ \textcolor{blue}{(+1.37)}\end{tabular}} \\ \bottomrule
\end{tabularx}
\end{table}

\begin{figure}[t]
  \centering
  \captionsetup[subfigure]{justification=centering}
  \begin{subfigure}{.45\linewidth}
    \centering
    \includegraphics[width=4.0cm]{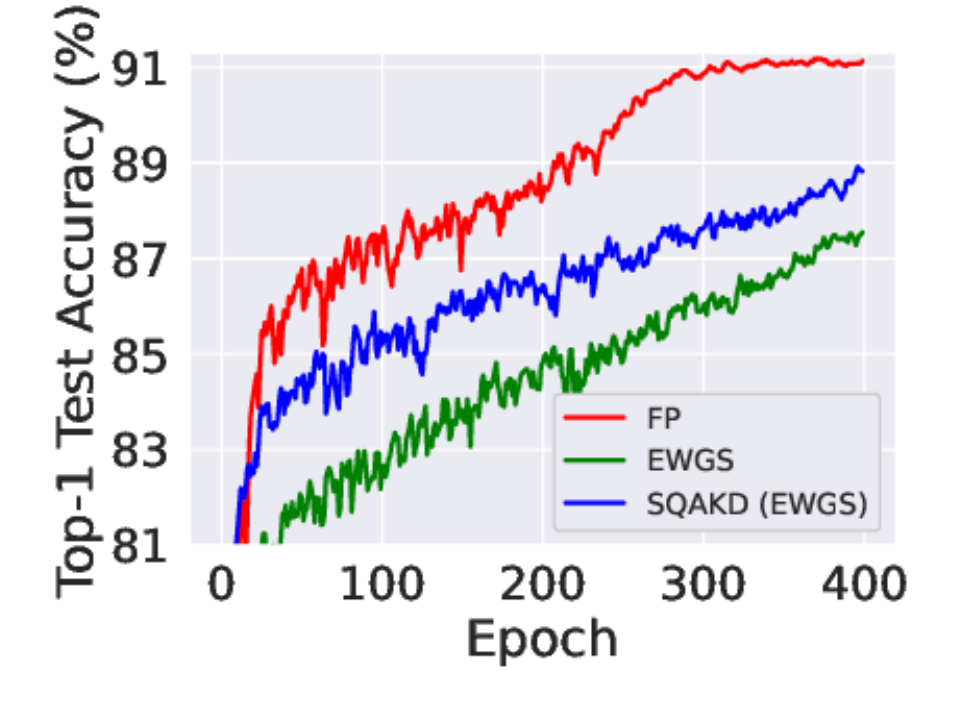}
    \vspace{-20pt}
    \caption{VGG-8 \\ (W1A1, CIFAR-10)}
    \label{fig:cifar10_vgg8_improvement_on_quantization}
  \end{subfigure}
  \begin{subfigure}{.45\linewidth}
    \centering
    \includegraphics[width=4.cm]{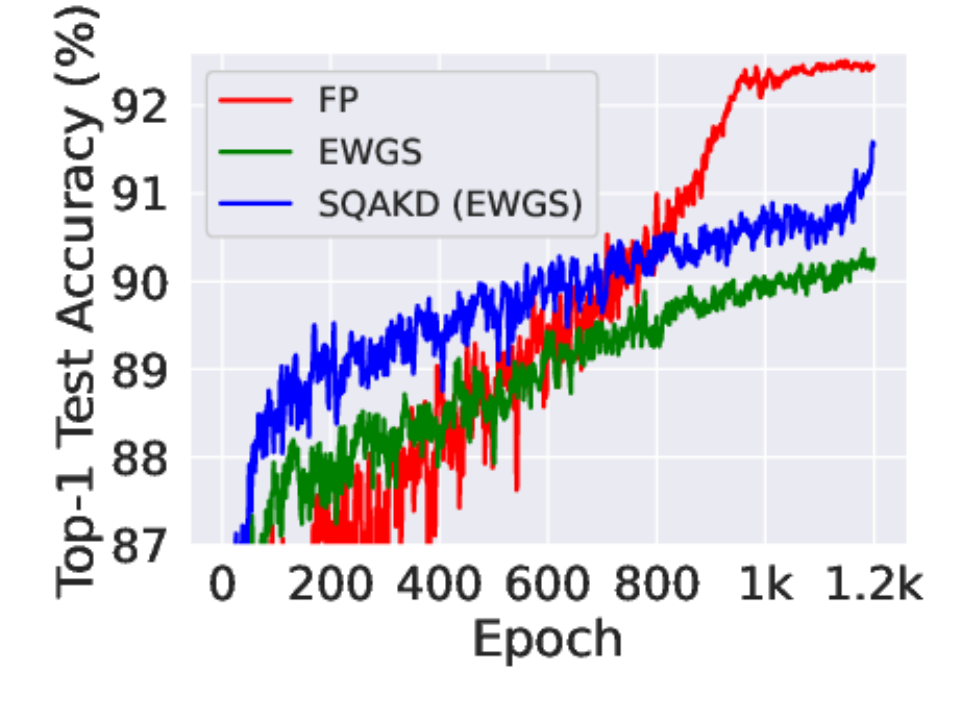}
    \vspace{-20pt}
    \caption{ResNet-20 \\ (W2A2, CIFAR-10)}
    \label{fig:cifar10_resnet20_improvement_on_quantization}
  \end{subfigure}
  \begin{subfigure}{.45\linewidth}
    \centering
    \includegraphics[width=4.cm]{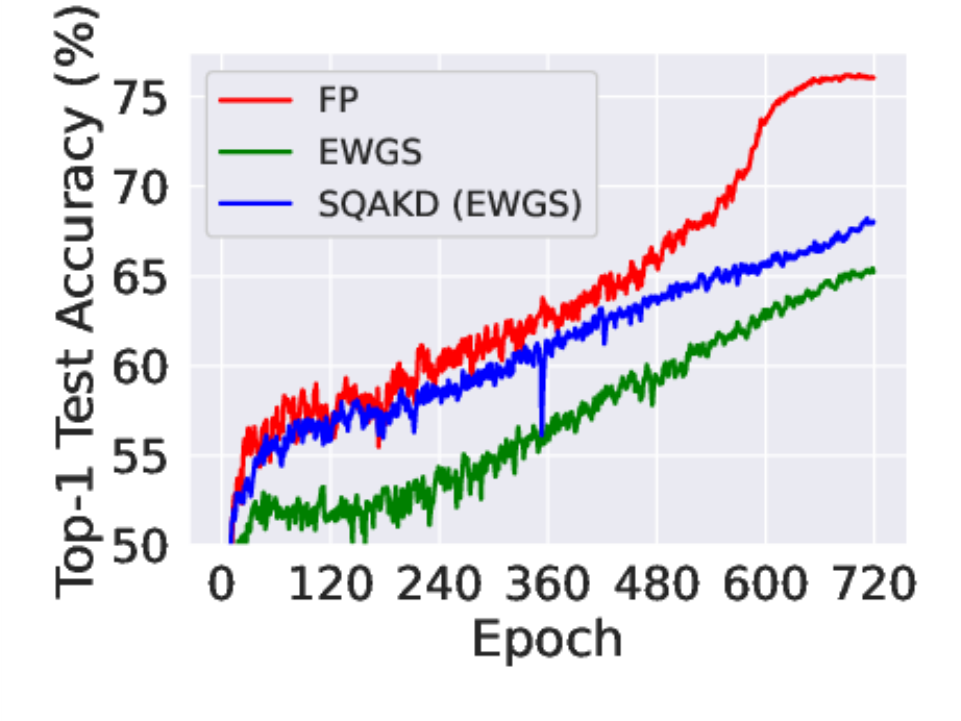}
    \vspace{-20pt}
    \caption{VGG-13 \\ (W1A1, CIFAR-100)}
    \label{fig:cifar100_vgg13_improvement_on_quantization}
  \end{subfigure}
  \begin{subfigure}{.45\linewidth}
    \centering
    \includegraphics[width=4.cm]{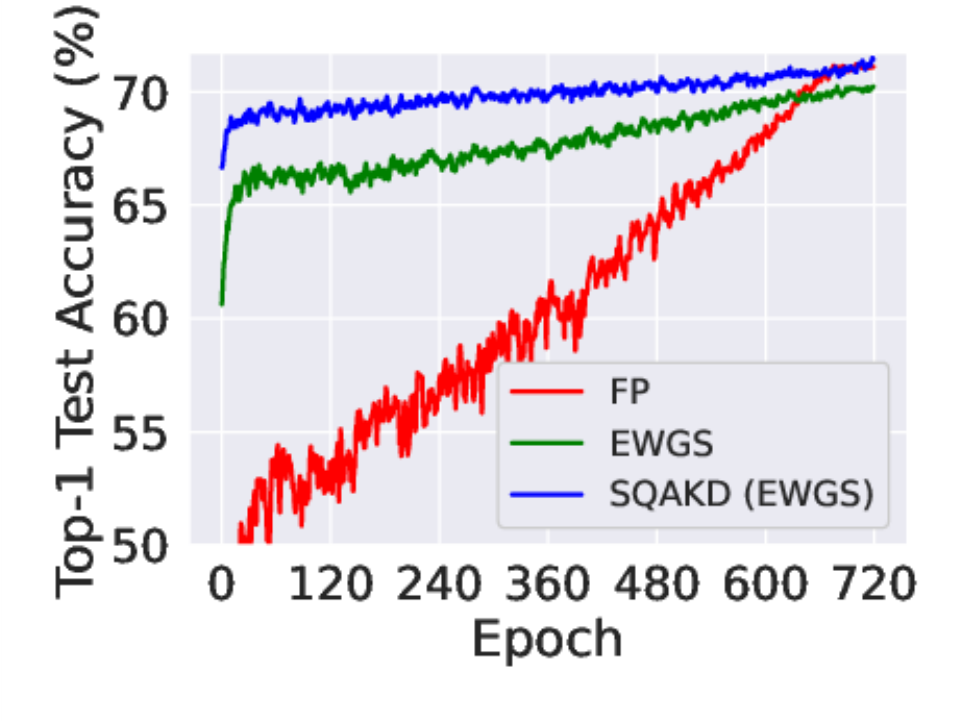}
    \vspace{-20pt}
    \caption{ResNet-32 \\ (W4A4, CIFAR-100)}
    \label{fig:cifar100_resnet32_improvement_on_quantization}
  \end{subfigure}
  \vspace{-5pt}
   \caption{Top-1 test accuracy evolution of full-precision (FP) and quantized models using the standalone EWES and SQAKD integrating EWGS, during training.} 
   \label{fig:cifar_improvement_on_quantization}
\end{figure}

\smallskip
\boldhdr{Results on CIFAR-10 and CIFAR-100}
Table~\ref{tab:cifar_improvement_on_quantization} presents the top-1 test accuracy of ResNet and VGG models on CIFAR-10 and CIFAR-100, where activations and weights are quantized to 1, 2, and 4 bits using 1) the standalone QAT method, i.e., EWGS, and 2) SQAKD that incorporates EWGS, i.e., SQAKD (EWGS).



SQAKD significantly improves the accuracy of EWGS in all bit quantization scenarios.
Specifically, on CIFAR-10, SQAKD improves EWGS by 0.36\% to 1.28\% on VGG-8 and 0.05\% to 0.39\% on ResNet-20; 
On CIFAR-100, the improvement is 1.26\% to 3.01\% on VGG-13 and 0.16\% to 1.15\% on ResNet-32.
Furthermore, SQAKD produces quantized models with higher accuracy than full-precision models. 
For example, on ResNet-32 with CIFAR-100, a 4-bit model trained with SQAKD outperforms the full-precision model by 0.32\%.
Such small and accurate models are useful for many real-world applications for edge deployment.
In comparison, the standalone EWGS leads to different levels of accuracy drop from 0.18\% to 6.16\% in all quantization scenarios.
Figure~\ref{fig:cifar_improvement_on_quantization} illustrates the top-1 test accuracy evolution for full-precision and quantized models in each epoch during training on CIFAR-10 and CIFAR-100. 
Compared to the standalone EWGS, SQAKD significantly accelerates the convergence speed of 1-bit VGG-8 and 2-bit ResNet-20 with CIFAR-10, as well as 1-bit VGG-13 and 4-bit ResNet-32 with CIFAR-100.
These results confirm that SQAKD improves EWGS in both converge speed and final accuracy.

\begin{table}[t]
\scriptsize
\centering
\caption{Top-1 and top-5 test accuracy (\%) of MobileNet-V2, ShuffleNet-V2, and SqueezeNet on Tiny-ImageNet.}
\vspace{-10pt}
\label{tab:tiny_imagenet_improvement_on_quantization_small_models}
\setlength{\tabcolsep}{0.5pt}
\begin{tabularx}{0.48\textwidth}{@{}lllll@{}}
\toprule
Model                          & \begin{tabular}[c]{@{}l@{}}Bit-\\width\end{tabular}             & Method                  & Top-1 Acc.               & Top-5 Acc.              \\ \midrule
\multirow{7}{*}{\begin{tabular}[c]{@{}l@{}}Mobile\\Net-V2\end{tabular}}  & FP                    & -                       & 58.07                    & 80.97                   \\
                               & \multirow{2}{*}{W3A3} & PACT                    & 47.77                    & 73.44                   \\
                               &                       & \textbf{SQAKD(PACT)}   & \textbf{52.73 \textcolor{blue}{(+4.96)}}   & \textbf{77.68 \textcolor{blue}{(+4.24)}}  \\
                               & \multirow{2}{*}{W4A4} & PACT                    & 50.33                    & 75.08                   \\
                               &                       & \textbf{SQAKD(PACT)}   & \textbf{57.14 \textcolor{blue}{(+6.81)}}   & \textbf{80.61 \textcolor{blue}{(+5.53)}}  \\
                               & \multirow{2}{*}{W8A8} & DoReFa                  & 56.26                    & 79.64                   \\
                               &                       & \textbf{SQAKD(DoReFa)} & \textbf{58.13 \textcolor{blue}{(+1.87)}}   & \textbf{81.3 \textcolor{blue}{(+1.66)}}   \\ \midrule
\multirow{5}{*}{\begin{tabular}[c]{@{}l@{}}Shuffle\\Net-V2\end{tabular}} & FP                    & -                       & 49.91                    & 76.05                   \\
                               & \multirow{2}{*}{W4A4} & PACT                    & 27.09                    & 52.54                   \\
                               &                       & \textbf{SQAKD(PACT)}   & \textbf{41.11  \textcolor{blue}{(+14.02)}} & \textbf{68.4 \textcolor{blue}{(+15.86)}}  \\
                               & \multirow{2}{*}{W8A8} & DoReFa                  & 45.96                    & 71.93                   \\
                               &                       & \textbf{SQAKD(DoReFa)} & \textbf{47.33 \textcolor{blue}{(+1.37)}}   & \textbf{73.85 \textcolor{blue}{(+1.92)}}  \\ \midrule
\multirow{5}{*}{\begin{tabular}[c]{@{}l@{}}Squee\\zeNet\end{tabular}}    & FP                    & -                       & 51.49                    & 76.02                   \\
                               & \multirow{2}{*}{W4A4} & LSQ                     & 35.37                    & 62.75                   \\
                               &                       & \textbf{SQAKD(LSQ)}    & \textbf{47.40  \textcolor{blue}{(+12.03)}} & \textbf{73.18 \textcolor{blue}{(+10.43)}} \\
                               & \multirow{2}{*}{W8A8} & DoReFa                  & 42.66                    & 69.25                   \\
                               &                       & \textbf{SQAKD(DoReFa)} & \textbf{46.62  \textcolor{blue}{(+3.96)}}  & \textbf{73.02 \textcolor{blue}{(+3.77)}}  \\ \bottomrule
\end{tabularx}
\end{table}

\smallskip
\boldhdr{Results on Tiny-ImageNet}
As shown in Table~\ref{tab:tiny_imagenet_improvement_on_quantization_large_models}, SQAKD consistently improves the top-1 accuracy of various QAT methods, including PACT, LSQ, and DoReFa, by a large margin for 3, 4, and 8-bit quantization on Tiny-ImageNet. 
For example, on 3-bit ResNet-18, SQAKD improves PACT by 4.31\%.
We observe that SQAKD achieves a more substantial improvement on low-bit quantization than on high-bit quantization.
For example, on ResNet-18, the improvements achieved by SQAKD using LSQ exhibit an escalating trend, measured at 0.88\%, 1.24\%, and 3.22\% respectively, as the bit width decreases from 8 to 4 and further to 3 bits.
Similar trends are observed in DoReFa with ResNet-18 and PACT with VGG-11.
This is because low-bit quantization causes more significant information loss, leading to reduced accuracy.
The distillation process in SQAKD mitigates this by leveraging insights from the teacher to guide the low-bit weight gradients.

SQAKD can also effectively quantize models (MobileNet, ShuffleNet, and SqueezeNet) that are already compact for use on resource-constrained edge devices, as shown in Table~\ref{tab:tiny_imagenet_improvement_on_quantization_small_models}. 
For example, on 4-bit quantization, for the top-1 and top-5 test accuracy, SQAKD improves 1) PACT by 14.02\% and 15.86\%, respectively, on ShuffleNet-V2; and 2) LSQ by 12.03\% and 10.43\%, respectively, on SqueezeNet.
On MobileNet-V2, for 8-bit quantization, SQAKD enables this already lightweight model to outperform the full-precision model by 0.06\% in top-1 and 0.33\% in top-5 accuracy;
and with 4-bit quantization, the accuracy drop achieved by SQAKD is within 1\%, whereas PACT causes a significant accuracy drop of 7.74\%.   
\subsection{Comparison with SOTA KD Methods}
\label{sec:comparison_with_KD}


\begin{table}[t]
\scriptsize
\centering
\caption{Top-1 test accuracy of KD in QAT, using EWGS as the quantizer. \textcolor{blue}{Blue} and \textcolor{red}{red} numbers inside the parenthesis denote the increase and decrease compared to the standalone EWGS, respectively.}
\vspace{-10pt}
\label{tab:comparison_with_KD}
\setlength{\tabcolsep}{5.0pt}
\begin{tabularx}{0.48\textwidth}{@{}lll@{}}
\toprule
Dataset       & CIFAR-10                                                        & CIFAR-100                                                    \\ \midrule
Model         & \begin{tabular}[c]{@{}l@{}}ResNet-20\\ (FP: 92.58)\end{tabular} & \begin{tabular}[c]{@{}l@{}}VGG-13\\ (FP: 76.36)\end{tabular} \\ \midrule
Bit-width           & W2A2                                                            & W1A1                                                         \\ 
EWGS~\citep{lee2021network}          & 91.41                                                           & 65.55                                                        \\
CRD~\citep{tian2019contrastive}    & 91.36 \textcolor{red}{(-0.05)}                                                   & 68.47 \textcolor{blue}{(+2.92)}                                                \\
AT~\citep{zagoruyko2016paying}     & 89.99 \textcolor{red}{(-1.42)}                                                   & 67.56 \textcolor{blue}{(+2.01)}                                                \\
NST~\citep{huang2017like}    & 89.07 \textcolor{red}{(-2.34)}                                                   & 66.04 \textcolor{blue}{(+0.49)}                                                \\
SP~\citep{tung2019similarity}     & 90.92 \textcolor{red}{(-0.49)}                                                   & 51.47 \textcolor{red}{(-14.08)}                                               \\
RKD~\citep{park2019relational}    & 90.91 \textcolor{red}{(-0.50)}                                                   & 67.23 \textcolor{blue}{(+1.68)}                                                \\
FitNet~\citep{romero2014fitnets} & 90.21 \textcolor{red}{(-1.20)}                                                   & 64.25 \textcolor{red}{(-1.30)}                                                \\
CC~\citep{peng2019correlation}     & 91.31 \textcolor{red}{(-0.10)}                                                   & 65.60 \textcolor{blue}{(+0.05)}                                                \\
VID~\citep{ahn2019variational}    & 88.40 \textcolor{red}{(-3.01)}                                                   & 65.88 \textcolor{blue}{(+0.33)}                                                \\
FSP~\citep{yim2017gift}    & 91.44 \textcolor{blue}{(+0.03)}                                                   & 65.28 \textcolor{red}{(-0.27)}                                                \\
FT~\citep{kim2018paraphrasing}     & 90.08 \textcolor{red}{(-1.33)}                                                   &   67.33 \textcolor{blue}{(+1.78)}                                                           \\
CKTF~\citep{zhao2023contrastive}   & 90.76 \textcolor{red}{(-0.65)}                                                   & 67.92 \textcolor{blue}{(+2.37)}                                                \\ 
\textbf{SQAKD}  & \textbf{91.80 \textcolor{blue}{(+0.39)}}                                          & \textbf{68.56 \textcolor{blue}{(+3.01)}}                                       \\
\bottomrule
\end{tabularx}
\end{table}


\begin{figure}[t]
  \centering
  \begin{subfigure}{.32\linewidth}
    \centering
    \includegraphics[width=2.85cm]{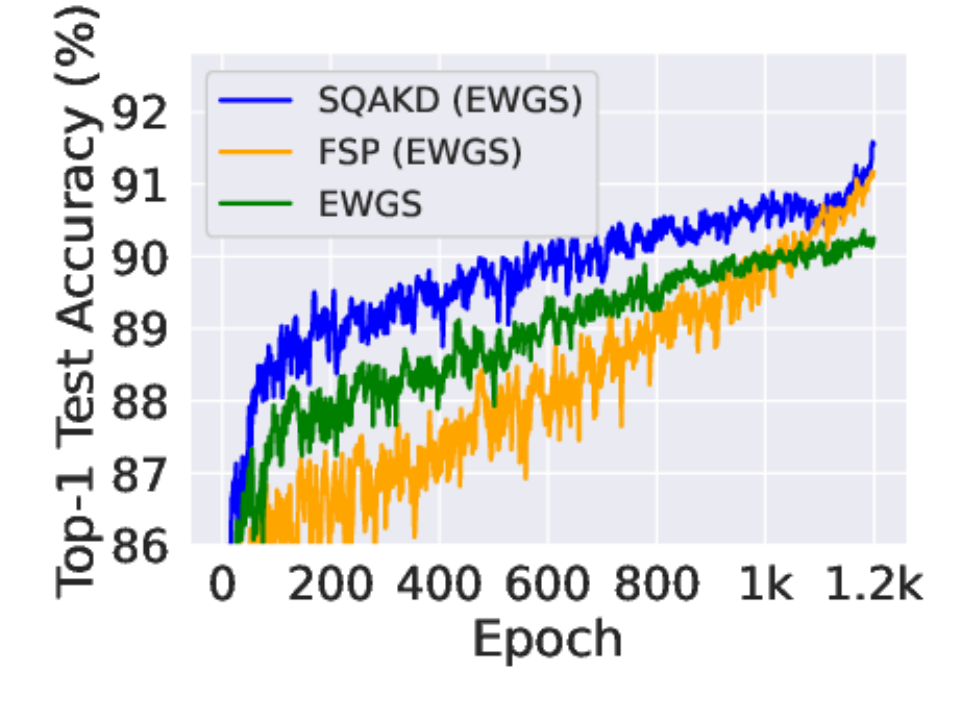}
    \vspace{-20pt}
    \caption{Test Accuracy}
    \label{fig:comparison_with_KD_accuracy}
  \end{subfigure}
  \begin{subfigure}{.32\linewidth}
    \centering
    \includegraphics[width=2.9cm]{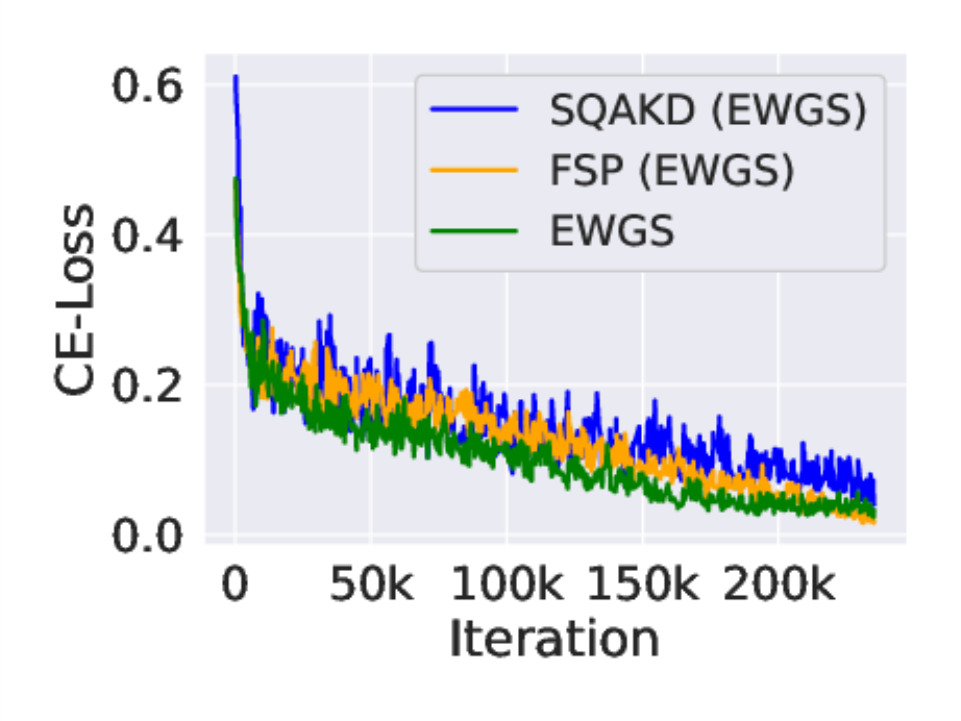}
    \vspace{-20pt}
    \caption{CE-Loss}
    \label{fig:comparison_with_KD_ce_loss}
  \end{subfigure}
  \begin{subfigure}{.32\linewidth}
    \centering
    \includegraphics[width=2.9cm]{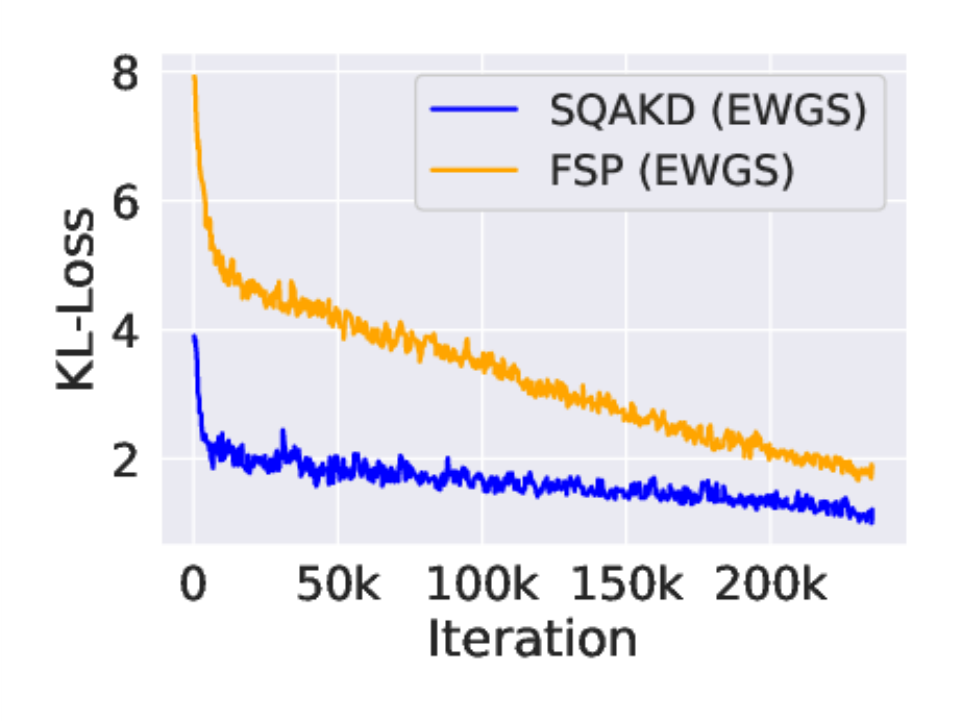}
    \vspace{-20pt}
    \caption{KL-Loss}
     \label{fig:comparison_with_KD_kl_loss}
  \end{subfigure}
  \vspace{-7pt}
  \caption{(a) Top-1 test accuracy, (b) CE-Loss, and (c) KL-Loss of EWGS, FSP, and SQAKD in each epoch/iteration during training on 2-bit ResNet-20 with CIFAR-10.}
   \label{fig:comparison_with_KD}
\end{figure}

Table~\ref{tab:comparison_with_KD} presents the top-1 test accuracy of the proposed SQAKD and 11 existing KD methods in the context of quantization (EWGS) for 2-bit ResNet-20 on CIFAR-10 and 1-bit VGG-13 on CIFAR-100.
We find out, in the context of quantization, the 11 existing KD methods cannot converge without the supervision from ground-truth labels.
So we compare the \textit{supervised} existing KD works for quantization with the proposed \textit{unsupervised} SQAKD in Table~\ref{tab:comparison_with_KD}.
%
The results show that \textit{unsupervised} SQAKD outperforms the \textit{supervised} KD methods by 0.36\% to 3.4\% on CIFAR-10 and 0.09\% to 17.09\% on CIFAR-100.



Compared to the standalone EWGS, SQAKD is 0.39\% better on CIFAR-10 and 3.01\% better on CIFAR-100.
In comparison, none of the existing KD works can consistently improve EWGS on both CIFAR-10 and CIFAR-100, and none can outperform SQAKD.

Also, as observed, compared to EWGS, SQAKD performs better on CIFAR-100 than on CIFAR-10 (3.01\% v.s. 0.39\%). 
This could be because CIFAR-100 is more complicated than CIFAR-10 (with more classes and data), resulting in more lost information and a higher accuracy drop during quantization. SQAKD is good at compensating for the lost information and recovering the accuracy by transferring knowledge from the full-precision teacher.

We provide a detailed comparison of SQAKD to FSP with EWGS as the quantizer, as FSP is the second-best in accuracy (shown in Table~\ref{tab:comparison_with_KD}). 
Figure~\ref{fig:comparison_with_KD_accuracy} shows that on 2-bit ResNet-20 with CIFAR-10, SQAKD converges much faster than the standalone EWGS and FSP, whereas FSP decreases the convergence speed of the standalone EWGS.
%
Figures~\ref{fig:comparison_with_KD_ce_loss} and \ref{fig:comparison_with_KD_kl_loss} show that, for CE-Loss, \textit{unsupervised} SQAKD achieves a comparable performance to \textit{supervised} EWGS and \textit{supervised} FSP; 
for KL-Loss, SQAKD converges much faster and achieves a lower final value than FSP.
This validates that SQAKD is sufficient for minimizing both CE-Loss and KL-Loss, leading to faster and better distillation.

The above results confirm that, in the context of quantization, \textit{self-supervised} SQAKD outperforms existing \textit{supervised} KD works in both convergence speed and final accuracy.

\subsection{Comparison with SOTA Methods Applying both QAT and KD}
\label{sec:comparison_with_quantization_KD_methods}


\begin{table}[t]
\scriptsize
\centering
\caption{Comparison of SQAKD with KD-integrated QAT methods.
Accuracy drop is measured by comparing the top-1 test accuracy of low-bit models with their full-precision counterparts. 
Negative values indicate an accuracy drop, while positive values denote an increase.}
\vspace{-10pt}
\label{tab:comparison_with_kd_and_quantization}
\begin{tabular}{@{}l|c|ccc@{}}
\toprule
Dataset        & CIFAR-10       & \multicolumn{3}{c}{CIFAR-100}                                         \\ \midrule
Model          & ResNet-20      & \multicolumn{1}{c|}{ResNet-32}      & \multicolumn{2}{c}{AlexNet}     \\
Bit-width      & W2A2           & \multicolumn{1}{c|}{W2A2}           & W2A2           & W4A4           \\
QKD            & -1.10          & \multicolumn{1}{c|}{-4.40}          & -              & -              \\
CMT-KD         & -              & \multicolumn{1}{c|}{-}              & -0.30          & -              \\
SPEQ           & -0.70          & \multicolumn{1}{c|}{\textbf{-}}     & -              & -              \\
PTG   & -              & \multicolumn{1}{c|}{-}              & +0.80          & +0.40          \\
\textbf{SQAKD} & \textbf{-0.66} & \multicolumn{1}{c|}{\textbf{-1.34}} & \textbf{+1.04} & \textbf{+2.29} \\ \bottomrule
\end{tabular}
\end{table}

Table~\ref{tab:comparison_with_kd_and_quantization} shows the comparison between QAKD and \kai{SOTA KD-integrated QAT methods that utilize both quantization and KD}, on CIFAR-10 and CIFAR-100. 
We measure the accuracy drop by comparing the top-1 test accuracy of the low-bit model with that of its corresponding full-precision model. 
\kai{The results of the related KD-integrated QAT works are directly obtained from their original papers, and the unavailable data is marked with `-'.}


SQAKD consistently outperforms all the baselines in all cases.
Specifically, for 2-bit ResNet models (ResNet-20 on CIFAR-10 and ResNet-32 on CIFAR-100), all the methods cause an accuracy drop while SQAKD achieves the smallest drop, lower than baselines by 0.04\% to 3.06\%.
For AlexNet on CIFAR-100, only SQAKD and PTG achieve higher accuracy than their full-precision model whereas the improvement achieved by SQAKD outperforms those of PTG, e.g., 1.04\% v.s. 0.8\% for 2-bit quantization and 2.29\% v.s. 0.4\% for 4-bit quantization.
In contrast, CMT-KD results in a 0.3\% accuracy degradation.

These baseline methods require a substantial amount of labeled data, intricate hyper-parameter tuning to balance the different loss terms, e.g., three parameters required in the loss function of CMT-KD, and complex training procedures, e.g., QKD with three-phase training. 
In contrast, SQAKD is self-supervised, does not require hyperparameter adjustments to balance the loss terms, and employs a simple-yet-effective single-phase training process.
These characteristics make SQAKD a more practical and efficient approach compared to the SOTA works.

\subsection{Inference Speedup}
\label{sec:inference_speedup}

\begin{table}[t]
\scriptsize
\centering
\caption{Inference on Jetson Nano with Tiny-ImageNet.}
\label{tab:inference_speedup}
\vspace{-10pt}
\setlength{\tabcolsep}{2.5pt}
\begin{tabularx}{0.48\textwidth}{@{}lcccc@{}}
\toprule
Model                           & Bit-width & Throughput (fps) & \begin{tabular}[c]{@{}c@{}}Inference \\ Time (s)\end{tabular} & Speedup \\ \midrule
\multirow{2}{*}{ResNet-18}      & FP32      & 1.8688                                                      & 0.5351                                                        & -       \\
                                & INT8      & 5.7925                                                      & 0.1726                                                        & 3.10$\times$    \\ \midrule
\multirow{2}{*}{MobileNet-V2}   & FP32      & 3.0905                                                      & 0.3236                                                        & -       \\
                                & INT8      & 9.4004                                                      & 0.1064                                                        & 3.04$\times$    \\ \midrule
\multirow{2}{*}{ShuffleNet-V2}  & FP32      & 12.6402                                                     & 0.0791                                                        & -       \\
                                & INT8      & 36.7817                                                     & 0.0272                                                        & 2.91$\times$    \\ \midrule
\multirow{2}{*}{SqueezeNet1\_0} & FP32      & 2.8105                                                      & 0.3558                                                        & -       \\
                                & INT8      & 8.7907                                                      & 0.1138                                                        & 3.13$\times$    \\ \bottomrule
\end{tabularx}
\end{table}

The reduction in model complexity in bit width that SQAKD achieves does translate to real speedup in model inference. 
We conduct inference experiments using the PyTorch framework and NVIDIA TensorRT on Jetson Nano, a widely used Internet-of-Things (IoT) platform. 
Table~\ref{tab:inference_speedup} shows that our method improves the inference speed by about 3$\times$ for 8-bit quantization using various model architectures, including ResNet-18, MobileNet-V2, ShuffleNet-V2, and SqueezeNet, on Tiny-ImageNet. 

%% file: ablation.tex
\section{Ablation Study}
\label{sec:ablation}

\begin{table}[t]
\scriptsize
\centering
\setlength{\tabcolsep}{2.5pt}
\caption{Top-1 test accuracy of different quantization forward and backward combinations.} 
\label{tab:ablation_forward_backward}
\vspace{-10pt}
\begin{tabularx}{0.48\textwidth}{@{}lllll@{}}
\toprule
Model and Dataset                                                                                             & Method                        & Forward               & Backward & Acc.  \\ \midrule
\multirow{4}{*}{\begin{tabular}[c]{@{}l@{}}ShuffleNet-V2\\ (W4A4, \\ Tiny-ImageNet)\end{tabular}} & FP                            & -                     & -        & 49.91 \\ \cmidrule(l){2-5} 
                                                                                                  & PACT                          & PACT                  & STE      & 27.09 \\ \cmidrule(l){2-5} 
                                                                                                  & \multirow{2}{*}{SQAKD (PACT)} & \multirow{2}{*}{PACT} & STE      & 41.11 \\
                                                                                                  &                               &                       & EWGS     & \textbf{41.88} \\ \midrule
\multirow{4}{*}{\begin{tabular}[c]{@{}l@{}}ResNet-20\\ (W2A2,\\ CIFAR-10)\end{tabular}}           & FP                            & -                     & -        & 92.58 \\ \cmidrule(l){2-5} 
                                                                                                  & EWGS                          & EWGS                  & EWGS     & 91.41 \\ \cmidrule(l){2-5} 
                                                                                                  & \multirow{2}{*}{SQAKD (EWGS)} & \multirow{2}{*}{EWGS} & STE      & 91.70 \\
                                                                                                  &                               &                       & EWGS     & \textbf{91.80}  \\ \bottomrule
\end{tabularx}
\end{table}

\begin{figure}[t]
  \centering
  \captionsetup[subfigure]{justification=centering}
  \begin{subfigure}{.31\linewidth}
    \centering
    \includegraphics[width=2.8cm]{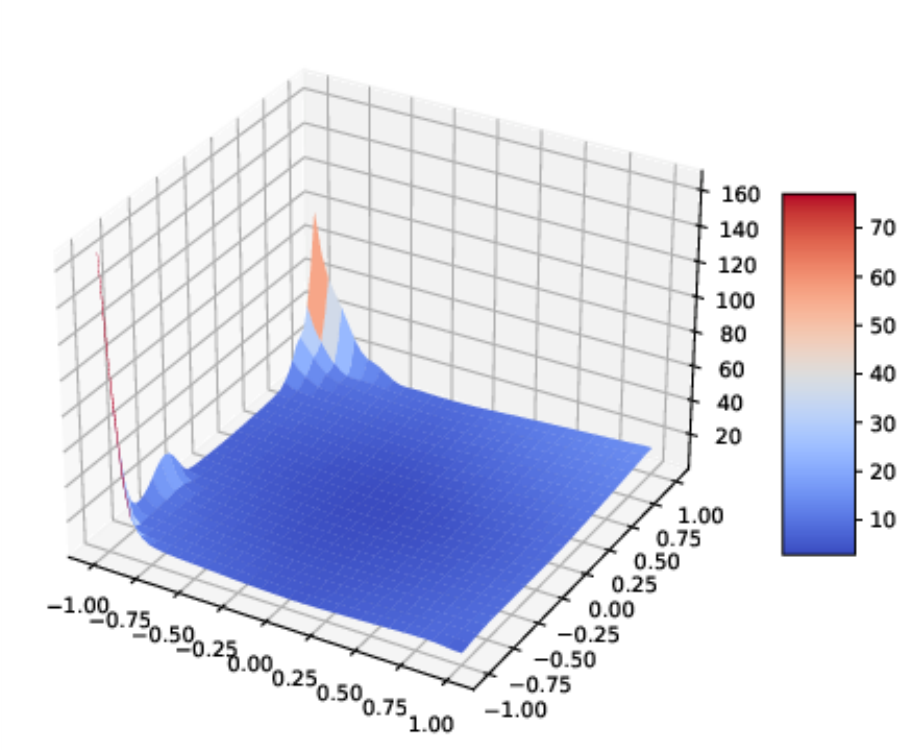}
    \vspace{-13pt}
    \caption{FP}
    \label{fig:cifar10_resnet20_fp_1200epochs_train_loss_3dsurface}
  \end{subfigure}
  \begin{subfigure}{.34\linewidth}
    \centering
    \includegraphics[width=2.8cm]{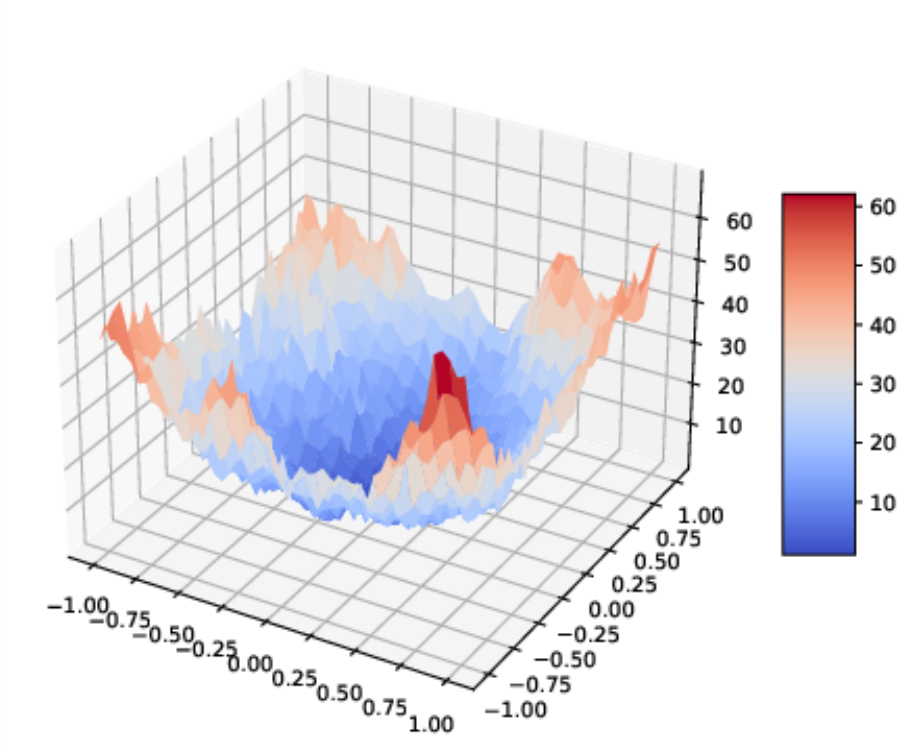}
    \vspace{-13pt}
    \caption{SQAKD (EWGS)}
    \label{fig:cifar10_resnet20_sqakd_1200epochs_train_loss_3dsurface}
  \end{subfigure}
  \begin{subfigure}{.32\linewidth}
    \centering
    \includegraphics[width=2.8cm]{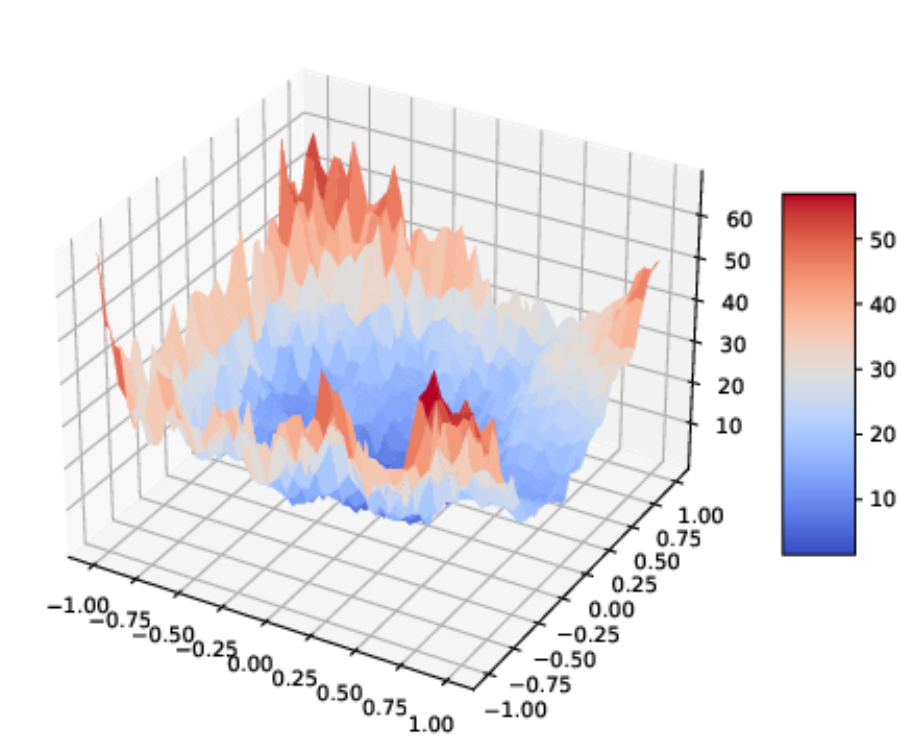}
    \vspace{-13pt}
    \caption{EWGS}
    \label{fig:cifar10_resnet20_ewgs_1200epochs_train_loss_3dsurface}
  \end{subfigure}
    \begin{subfigure}{.31\linewidth}
    \centering
    \includegraphics[width=2.8cm]{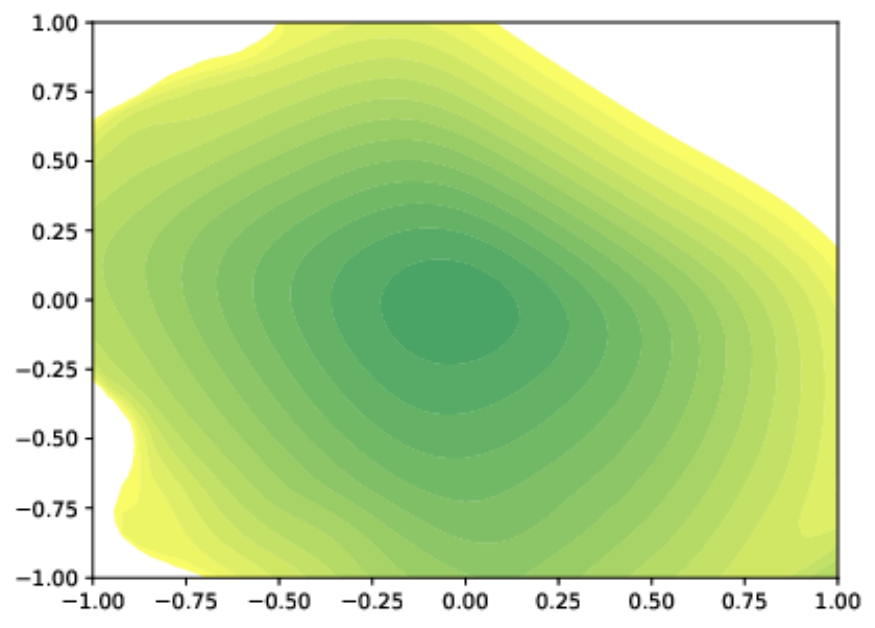}
    \vspace{-13pt}
    \caption{FP}
    \label{fig:cifar10_resnet20_fp_1200epochs_train_loss_2dcontourf}
  \end{subfigure}
  \begin{subfigure}{.34\linewidth}
    \centering
    \includegraphics[width=2.8cm]{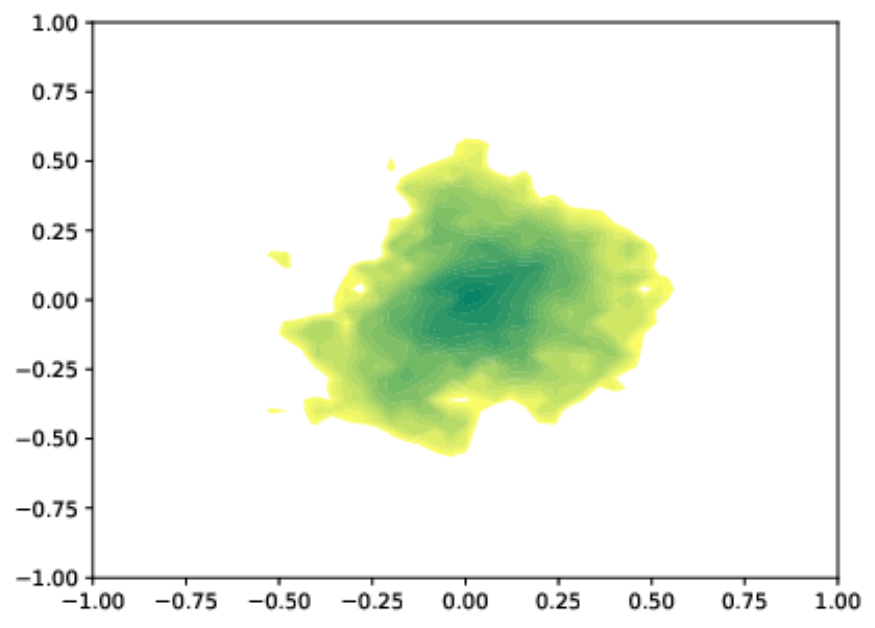}
    \vspace{-13pt}
    \caption{SQAKD (EWGS)}
    \label{fig:cifar10_resnet20_sqakd_1200epochs_train_loss_2dcontourf}
  \end{subfigure}
  \begin{subfigure}{.32\linewidth}
    \centering
    \includegraphics[width=2.8cm]{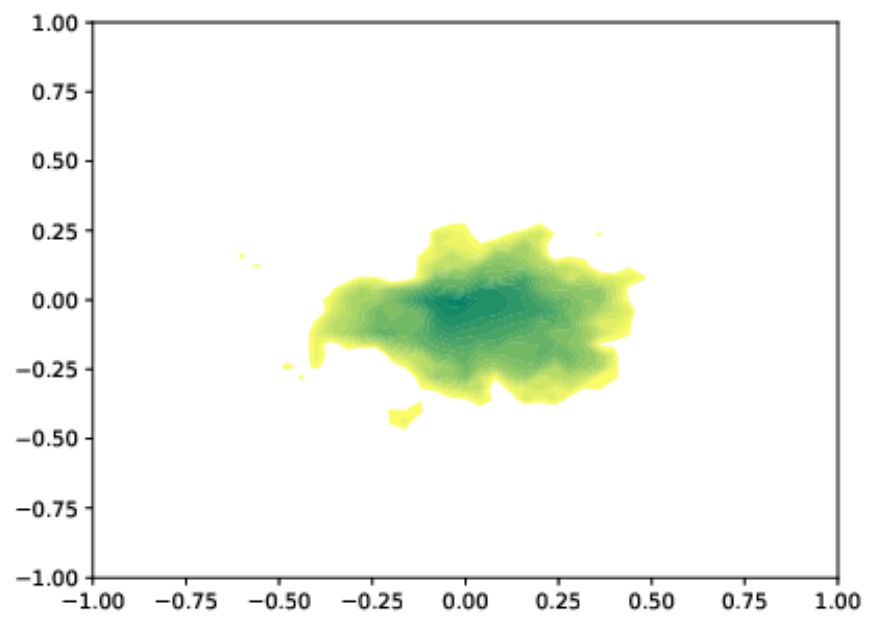}
    \vspace{-13pt}
    \caption{EWGS}
    \label{fig:cifar10_resnet20_ewgs_1200epochs_train_loss_2dcontourf}
  \end{subfigure}
 \vspace{-10pt}
  \caption{3D loss surface (a, b, c) and 2D contours (d, e, f) for full-precision and 2-bit ResNet-20 using SQAKD and EWGS on CIFAR-10.}
  \label{fig:ablation_loss_surface}
\end{figure}

\begin{figure}[t]
  \centering
  \captionsetup[subfigure]{justification=centering}
  \begin{subfigure}{.45\linewidth}
    \centering
    \includegraphics[width=4.2cm]{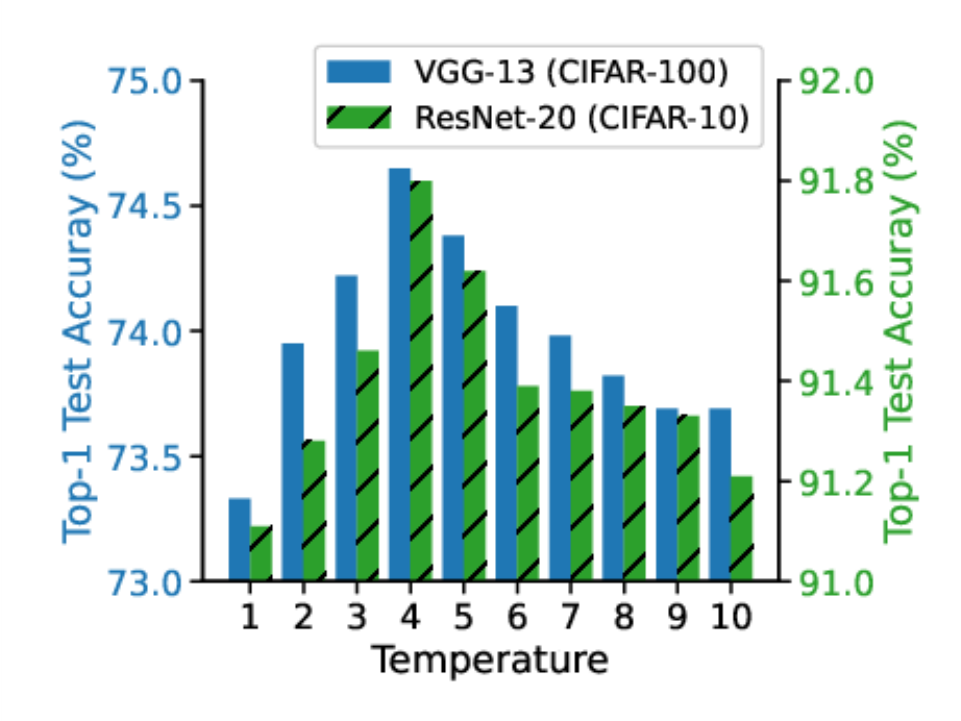}
    \vspace{-20pt}
    \caption{Temperature}
    \label{fig:ablation_temperature}
  \end{subfigure}
  \begin{subfigure}{.45\linewidth}
    \centering
    \includegraphics[width=4.2cm]{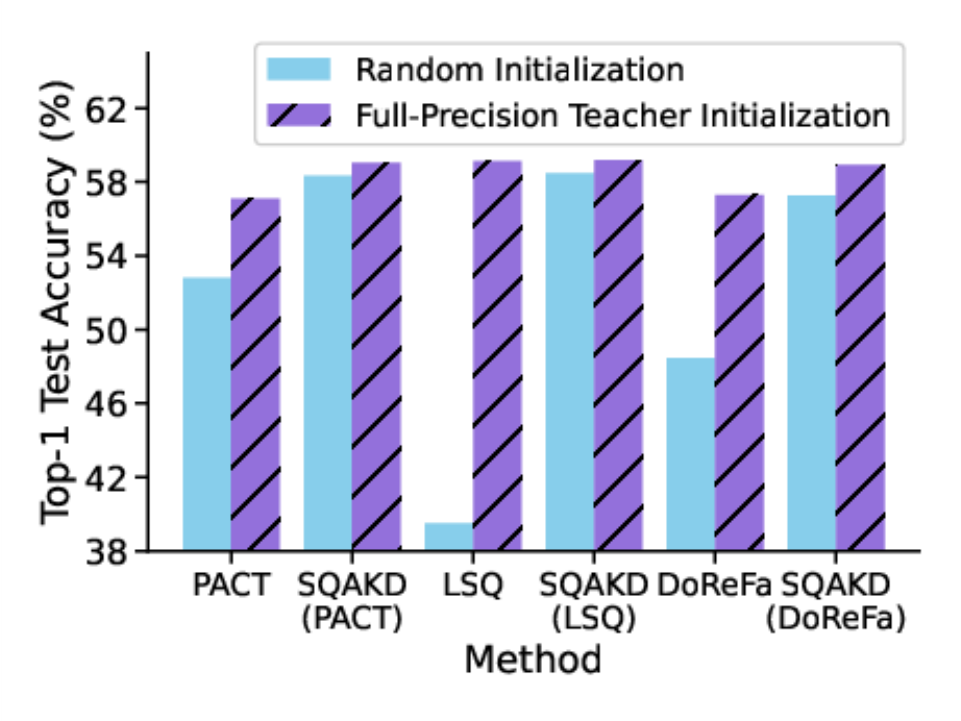}
    \vspace{-20pt}
    \caption{Initialization}
    \label{fig:ablation_initialization}
  \end{subfigure}
  \vspace{-7pt}
  \caption{Effect of (a) temperature (W2A2), and (b) initialization (W4A4, VGG-11, Tiny-ImageNet).}
\end{figure}

\smallskip
\boldhdr{Analysis of Loss Surface}
Figure~\ref{fig:ablation_loss_surface} compares the 3D loss surface and the corresponding 2D contour of the full-precision ResNet-20 and the 2-bit ResNet-20 trained using SQAKD and the standalone EWGS on CIFAR-10. 
SQAKD enables the quantized model to achieve a flatter and smoother loss surface compared to using the standalone EWGS.
To the best of our knowledge, we are the first to visualize and analyze the loss surface of a low-precision model achieved using QAT. 

\smallskip
\boldhdr{Flexibility for various forward and backward combinations}
To show the flexibility of our framework, we modularly integrate forward and backward techniques of SOTA QAT methods, like PACT and EWGS, into SQAKD.
Table~\ref{tab:ablation_forward_backward} shows that, on 4-bit ShuffleNet-V2 with Tiny-ImageNet, SQAKD enhances PACT by 14.02\% using STE backward and further improves it by 0.77\% using EWGS backward.
The success of the EWGS backward results from its integration of discretization error into gradient approximation.
On 2-bit ResNet-20 with CIFAR-10, SQAKD improves EWGS by 0.29\% and 0.39\% using STE and EWGS backward, respectively.
We are the first to decouple QAT's forward and backward techniques and to provide a detailed performance analysis.


\smallskip
\boldhdr{Effect of Temperature}
Temperature $\rho$ (in Eq.~\ref{equ:optimization_SQAKD}) softens the distribution, helping extract the teacher's dark knowledge.
We conducted an experimental investigation over the range $\rho \in [1,10]$ using VGG-13 on CIFAR-100 and ResNet-20 on CIFAR-10. 
Figure~\ref{fig:ablation_temperature} shows that $\rho=4$ yields the best performance.
These findings align with the empirical insights in contrastive knowledge transfer~\citep{tian2019contrastive, zhao2023contrastive}.


\smallskip
\boldhdr{Effect of Initialization}
Figure~\ref{fig:ablation_initialization} shows that, with the same setting, 
initialized either randomly or with the full-precision teacher, 
SQAKD increases the top-1 test accuracy of PACT, LSQ, and DoReFa by different levels (from 0.05\% to 18.97\%) for 4-bit VGG-11 on Tiny-ImageNet.
Furthermore, the full-precision teacher initialization outperforms the random initialization in all cases.

%% file: conclusion.tex
\section{Conclusion}

This paper proposes a Self-Supervised Quantization-Aware Knowledge Distillation (SQAKD) framework for model quantization in image classification.
SQAKD establishes stronger baselines and does not require labeled training data, potentially making QAT research more accessible.
It also provides a generalized agreement to existing QAT methods by unifying the optimization of their forward and backward dynamics.
SQAKD is hyperparameter-free and requires simple training procedures and low training costs, making it a highly practical and usable solution.
We quantitatively investigate and benchmark 11 KD methods in the context of QAT, providing a novel perspective on KD beyond its typical applications.
Our results confirm that SQAKD consistently outperforms the existing KD and QAT approaches by a large margin.

%


%% file: acknowledgments.tex
\section{Acknowledgments}
\kai{We thank the anonymous reviewers for their feedback. 
This work is supported by National Science Foundation awards CNS-2311026, SES-2231874, OAC-2126291 and CNS-1955593.}

%% file: supplement.tex
\onecolumn
\aistatstitle{Self-Supervised Quantization-Aware Knowledge Distillation: \\
Supplementary Materials}


\section{Experiment Setup}



\subsection{Models and Datasets}
We conducted extensive experiments to evaluate our proposed Self-Supervised Quantization-Aware Knowledge Distillation (SQAKD) framework on various image classification datasets, including
1) CIFAR-10~\citep{krizhevsky2009learning} which contains 50K (32$\times$32) RGB training images, belonging to 10 classes,
2) CIFAR-100~\citep{krizhevsky2009learning} which consists of 50K 32$\times$32 RGB training images with 100 classes, and 
3) Tiny-ImageNet~\citep{le2015tiny} containing 100K 64$\times$64 training images with 200 classes.

We utilized the widely used baseline architectures, ResNet~\citep{he2016identity} and VGG~\citep{eigen2015predicting}. 
We also select lightweight architectures, including MobileNet~\citep{sandler2018mobilenetv2}, ShuffleNet~\citep{zhang2018shufflenet}, and SqueezeNet~\citep{iandola2016squeezenet}.
We initialize the model weights using its corresponding pre-trained, full-precision counterparts.
Following the commonly used experimental settings~\citep{li2021mqbench,lee2021network, rastegari2016xnor,zhang2018lq}, we do not quantize the first convolutional layer and the last fully-connected layer.

\subsection{Implementation Details} 
We used four NVIDIA RTX 2080 GPUs to train the models.
The implementation details are shown in Table~\ref{tab:appedix_implementation_details}. 
On Tiny-ImageNet, the learning rate rises up to 5e-4 linearly in the first 2500 iterations and then decays to 0.0 via cosine annealing;
On CIFAR-10 and CIFAR-100, the learning rate starts with 1e-3 and 5e-4, respectively, and decays to 0.0 using a cosine annealing schedule.
For all training images, we apply basic data augmentation, including random cropping, horizontal flipping, and normalization.
Specifically, the training images are cropped to $224 \times 224$ for Tiny-ImageNet and $32 \times 32$ for CIFAR-10/CIFAR-100.

We conducted inference experiments on Jetson Nano using NVIDIA TensorRT~\citep{tensorrt}.
We first converted the model from PyTorch to ONNX, then transformed the ONNX representation into a TensorRT engine file, and finally deployed it using the TensorRT runtime.
The NVIDIA Jetson Nano is a compact computing platform designed for edge computing and artificial intelligence (AI) applications. 
Built on NVIDIA's advanced GPU architecture, the Jetson Nano provides computing capabilities in a small footprint, making it ideal for deploying AI models in embedded systems and low-power environments. 
NVIDIA TensorRT is a high-performance deep learning inference library, which supports deep learning applications in environments where computational resources are constrained.



\begin{table}[htp]
\small
\centering
\caption{Implementation details.}
\label{tab:appedix_implementation_details}
\vspace{-10pt}
\setlength{\tabcolsep}{6.5pt}
\begin{tabularx}{0.98\textwidth}{@{}lllllll@{}}
\toprule
Dataset                        & Model         & Initial learning rate & Training epochs & Batch size & Optimizer & Weight decay \\ \midrule
\multirow{2}{*}{CIFAR-10}      & VGG-8         & 1.00E-03              & 400             & 256        & Adam      & 1.00E-04     \\
                               & ResNet-20     & 1.00E-03              & 1200            & 256        & Adam      & 1.00E-04     \\ \midrule
\multirow{2}{*}{CIFAR-100}     & VGG-13        & 5.00E-04              & 720             & 64         & Adam      & 5.00E-04     \\
                               & ResNet-32     & 5.00E-04              & 720             & 64         & Adam      & 5.00E-04     \\ \midrule
\multirow{5}{*}{Tiny-ImageNet} & ResNet-18     & 5.00E-04              & 100             & 64         & SGD       & 5.00E-04     \\
                               & VGG-11        & 5.00E-04              & 100             & 32         & SGD       & 5.00E-04     \\
                               & MobileNet-V2  & 5.00E-04              & 100             & 32         & SGD       & 5.00E-04     \\
                               & ShuffleNet-V2 & 5.00E-04              & 100             & 64         & SGD       & 5.00E-04     \\
                               & SqueezeNet    & 5.00E-04              & 100             & 64         & SGD       & 5.00E-04     \\ \bottomrule
\end{tabularx}
\end{table}

\subsection{Hyper-parameters of baselines}
\kai{For a fair comparison of state-of-the-art (SOTA) QAT methods, we reran their original open-source code, achieving results consistent with those in the respective papers. 
Specifically, we used the original EWGS code~\citep{lee2021network} and the MQbench code~\citep{li2021mqbench} (a commonly used baseline code for QAT works) for PACT, LSQ, and DoReFa, as their original code is not available.
For the related 11 KD baselines, we implemented them using the CRD code~\citep{tian2019contrastive}, which is a widely used baseline code and has been confirmed with the authors of those KD works. 
We also strictly used the exactly same hyperparameters as those reported in CRD~\citep{tian2019contrastive} and the original papers of KD works.}
For example, we implemented SOTA KD methods in the context of QAT, and their loss functions are represented as: $L = L_{CE} + \lambda L_{Distill}$, where $L_{CE}$ is the cross-entropy loss with labels, $L_{Distill}$ is the distillation loss for transferring knowledge between the teacher and the student, and $\lambda$ controls the weights of the loss terms. 
Note that different KD methods have different forms of $L_{Distill}$, and $L_{Distill}$ can contain multiple loss terms.
For example, CKTF contains two loss terms in $L_{Distill}$, while VID and FSP set the number of loss terms the same as the number of intermediate layer pairs between the teacher and student.
The hyper-parameter settings $\lambda$ are the same as those used in the original papers, shown in Table~\ref{tab:hyperparameters_KD}. 


\begin{table}[htp]
\centering
\caption{Hyper-parameters of related KD methods}
\label{tab:hyperparameters_KD}
\vspace{-10pt}
\setlength{\tabcolsep}{9.0pt}
\begin{tabularx}{0.98\textwidth}{@{}llllllllllll@{}}
\toprule
Method    & CRD & AT   & NST & SP   & RKD & FitNet & CC   & VID & FSP & FT  & CKTF \\ \midrule
$\lambda$ & 0.8 & 1000 & 50  & 3000 & 1.0   & 100    & 0.02 & 1.0   & 50  & 200 & 0.8  \\ \bottomrule
\end{tabularx}
\end{table}

\section{Evaluation}

\subsection{Improvements on SOTA QAT Methods}

\begin{figure}[htp]
  \centering
  \begin{subfigure}{.24\linewidth}
    \centering
    \includegraphics[width=4.5cm]{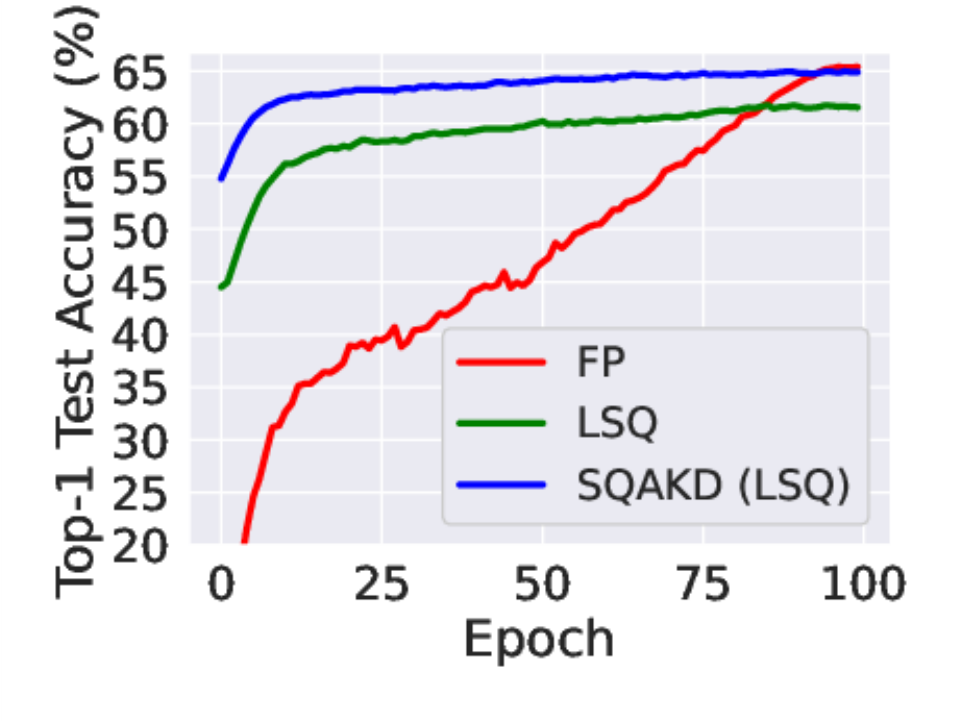}
    \caption{ResNet-18 (W3A3)}
  \end{subfigure}
  \begin{subfigure}{.24\linewidth}
    \centering
    \includegraphics[width=4.5cm]{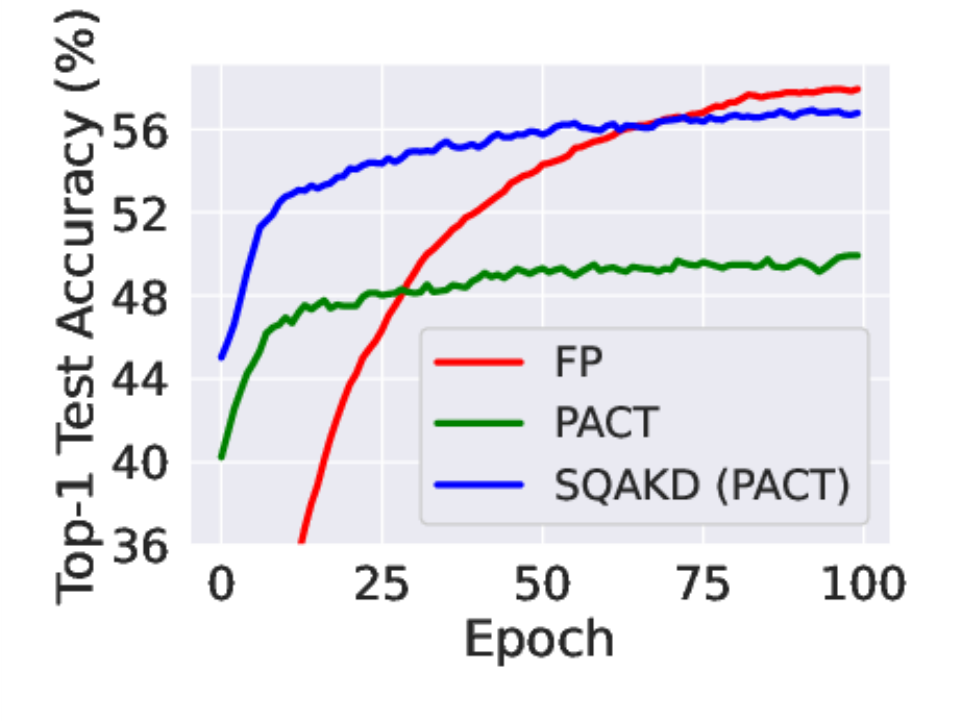}
    \caption{MobileNet-V2 (W4A4)}
  \end{subfigure}
  \begin{subfigure}{.24\linewidth}
    \centering
    \includegraphics[width=4.5cm]{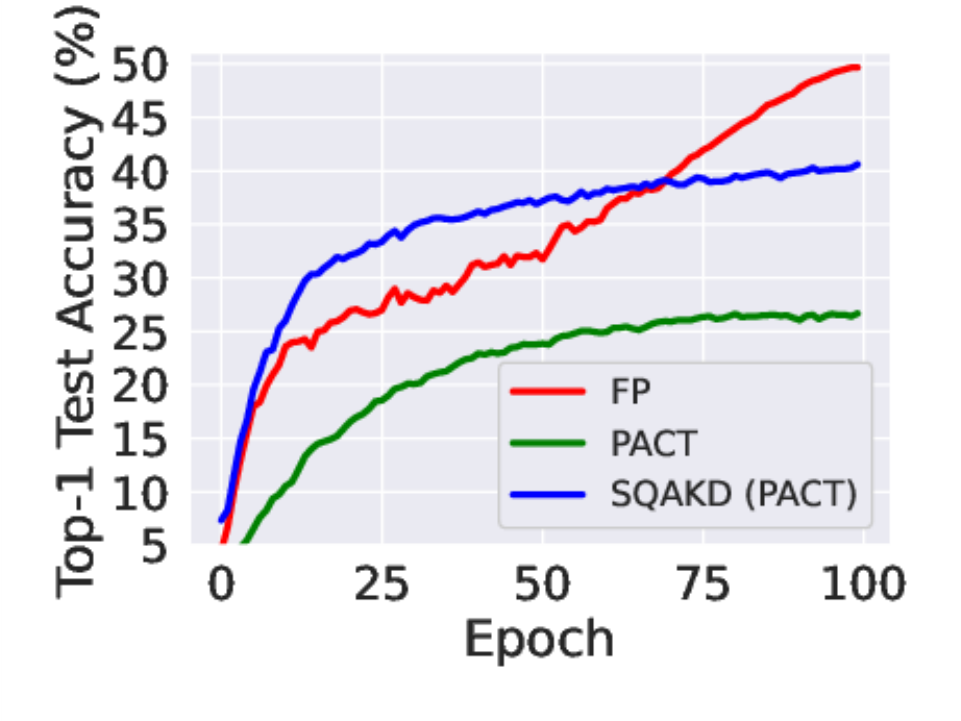}
    \caption{ShuffleNet-V2 (W4A4)}
  \end{subfigure}
  \begin{subfigure}{.24\linewidth}
    \centering
    \includegraphics[width=4.5cm]{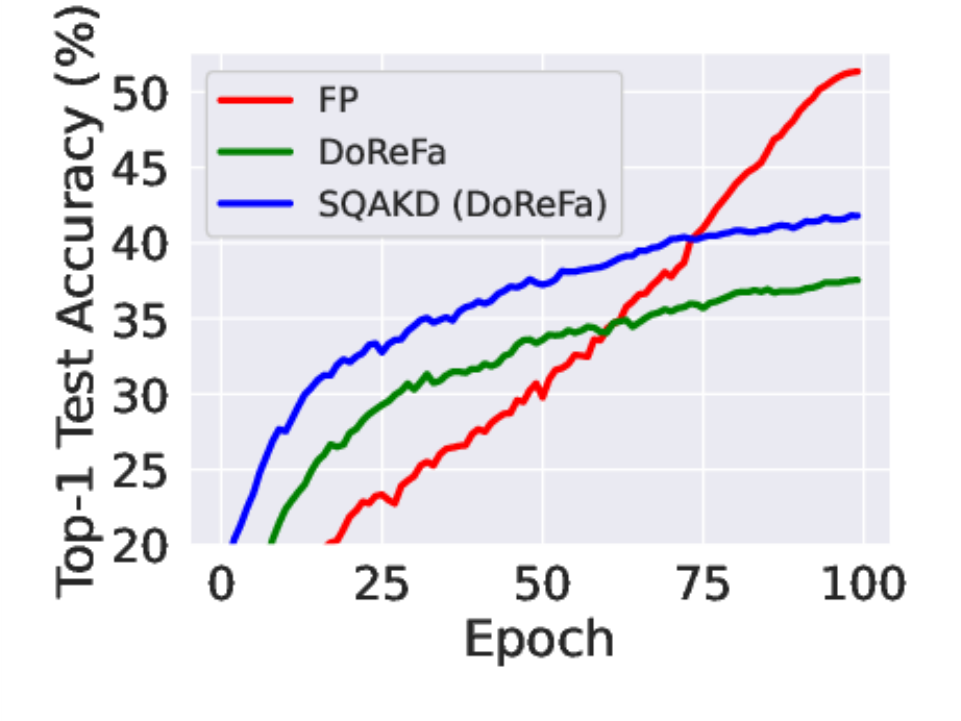}
    \caption{SqueezeNet (W4A4)}
  \end{subfigure}
\caption{Top-1 test accuracy evolution of full-precision models (FP), and quantized models using SQAKD and the standalone quantization methods, including LSQ, PACT, and DoReFa, in each epoch during training on Tiny-ImageNet.} 
\label{fig:tiny_imagenet_improvement_on_quantization}
\end{figure}

\smallskip
\boldhdr{Results on Tiny-ImageNet}
Figure~\ref{fig:tiny_imagenet_improvement_on_quantization} shows that SQAKD improves the convergence speed of all the baselines on ResNet-18, MobileNet-V2, ShuffleNet-V2 and SqueezeNet, with Tiny-ImageNet.
Furthermore, on MobileNet-V2, SQAKD enables the quantized student to converge much faster than the full-precision teacher, whereas the standalone QAT methods cannot achieve that.

\smallskip
\boldhdr{Results on CIFAR-100}
We observe that the accuracy improvement achieved by SQAKD is higher on VGG models than on ResNet models with the same level of quantization.
For example, for 1-bit quantization, on CIFAR-100, the accuracy improvement on VGG-13 is 3.01\%, which is much higher than 0.16\% on ResNet-32; and on CIFAR-10, the accuracy improvement of 1.28\% on VGG-8 outperforms that of 0.05\% on ResNet-20.
This might be because VGG models have a simpler, more explicit, and more structured hierarchy of features than ResNet models.
The hierarchical structure in VGG models, with sequential convolutional layers, helps the student to learn representations at different levels of abstraction and to recover the accuracy loss caused by quantization.
In comparison, the skip connections in ResNet architectures cause irregular, nonlinear pathways, complicating the knowledge transfer process.

\subsection{Comparison with SOTA KD Methods}
We observe that, with supervision by labels, the KD works that transfer structural knowledge of outputs perform better than those that transfer conditionally independent outputs.
For example, on CIFAR-100, with EWGS as the quantizer, SP~\citep{tung2019similarity} and FitNet~\citep{romero2014fitnets} underperform the standalone EWGS. 
In contrast, CRD~\citep{tian2019contrastive}, RKD~\citep{park2019relational}, and CKTF~\citep{zhao2023contrastive}, which capture the structural relations of intermediate representations (CKTF) or penultimate outputs (CRD and RKD), perform well.
This might be because, by capturing the correlations of high-order output dependencies from the full-precision model, it effectively restores the lost information resulting from quantization and guides the gradient updates of the low-bit model.
Although KD has been studied in the existing works, to the best of our knowledge, we are the first to study the performance of various KD methods in the context of quantization.


\section{Ablation study}

\subsection{Effect of Training Time}
\begin{figure}[htp]
    \centering
    \includegraphics[width=0.40\linewidth]{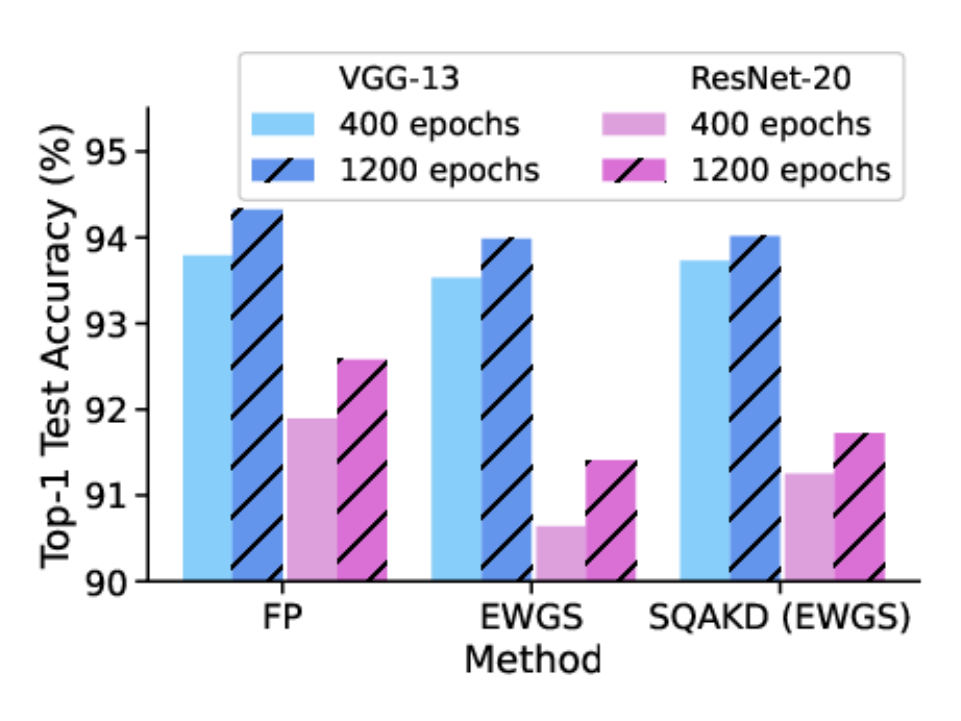}
    \vspace{-15pt}
    \caption{Effect of training Time (W2A2, CIFAR-10).}
    \label{fig:ablation_training_time}
    \vspace{-25pt}
\end{figure}

Motivated by the proposition that prolonged training increases test accuracy~\citep{he2019bag}, we investigated its applicability to QAT. 
Figure~\ref{fig:ablation_training_time} illustrates that for 2-bit VGG-13 and ResNet-20 on CIFAR-10, regardless of the approach --- full precision (FP), EWGS, and SQAKD, a 1200-epoch training span consistently outperforms 400 epochs, with accuracy gains from 0.28\% to 0.77\%.
Also, given equal training epochs, SQAKD yields consistent improvements over EWGS, from 0.2\% to 0.61\% for 400 epochs and 0.03\% to 0.31\% for 1200 epochs.

\begin{figure}[htp]
	\centering
        \begin{subfigure}{.45\linewidth}
		\centering
  		\includegraphics[width=5.5cm]{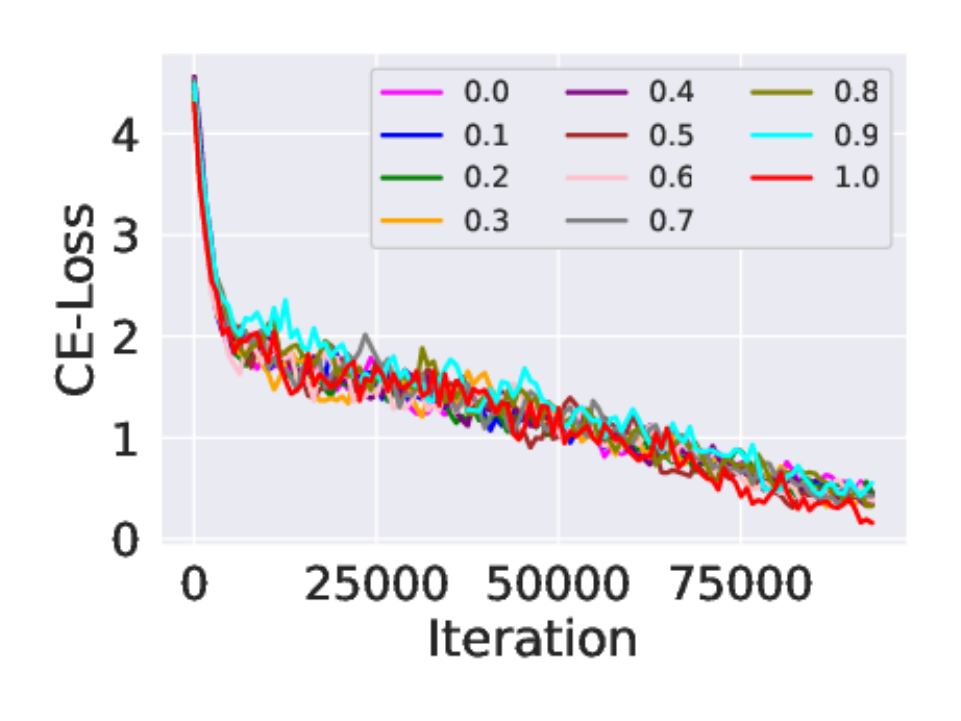}
            \vspace{-10pt}
		\caption{CE-Loss}
            \label{fig:loss_cifar100_vgg13_ours_ce}
	\end{subfigure}
        \begin{subfigure}{.45\linewidth}
		\centering
		\includegraphics[width=5.5cm]{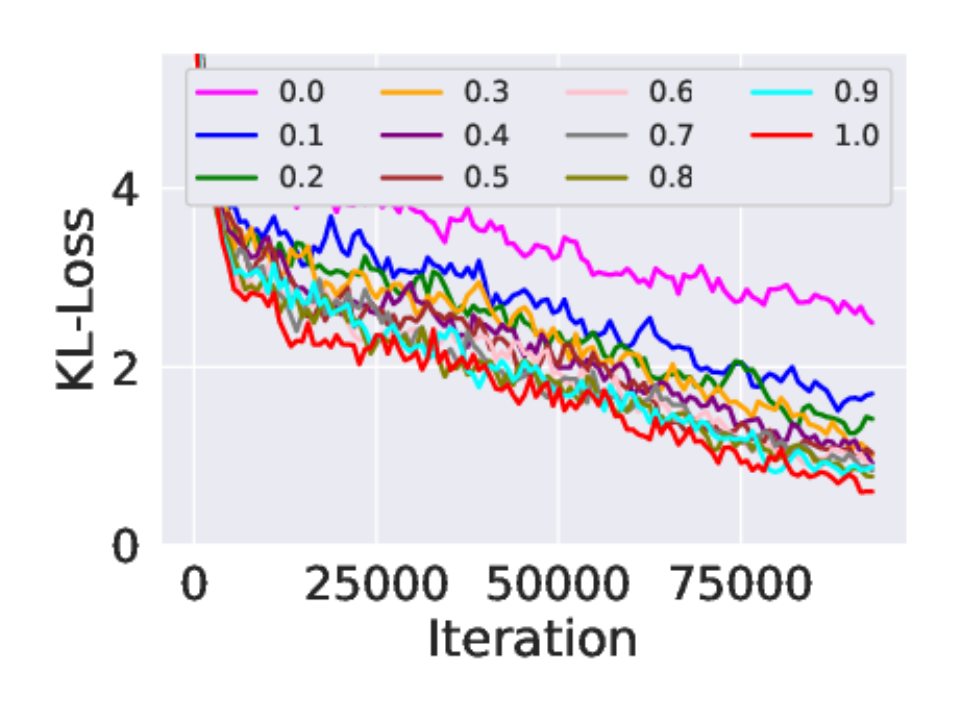}
		\vspace{-10pt}
            \caption{KL-Loss} 
            \label{fig:loss_cifar100_vgg13_ours_kl}
	\end{subfigure}
	\vspace{-6pt}
    \caption{The evolution of (a) CE-Loss and (b) KL-Loss during the training of 1-bit VGG-13 on CIFAR-100, with $\lambda \in \{0.0,0.1,…,1.0\}$. When $\lambda=0.0$, only CE-Loss is minimized; when $\lambda=1.0$, only KL-Loss is minimized.}
    \label{fig:loss_cifar100_vgg13}
	\vspace{-10pt}
\end{figure}

\kai{
\subsection{Analysis of CE-Loss and KL-Loss}
In addition to the three scenarios shown in Figure~\ref{fig:loss_analysis} where $\lambda \in \{0.0,0.5,1.0\}$ (Eq.~\ref{equ:optimization_ce_distill}), we extend our analysis by considering $\lambda \in \{0.0,0.1,0.2,0.3,…,0.9,1.0\}$ on 1-bit VGG-13 with CIFAR-100 to further investigate the relationship between the cross-entropy loss (CE-Loss) and the KL-divergence loss (KL-Loss).
Figure~\ref{fig:loss_cifar100_vgg13} shows that solely minimizing KL-Loss ($\lambda=1.0$) effectively leads to the concurrent minimization of both CE-Loss and KL-Loss.
These results validate that solely minimizing KL-Loss is sufficient to achieve optimal gradient updates in the quantized network.
}

\subsection{Analysis of Loss Surface}

\begin{figure}[htp]
  \centering
  \captionsetup[subfigure]{justification=centering}
  \begin{subfigure}{.32\linewidth}
    \centering
    \includegraphics[width=4.6cm]{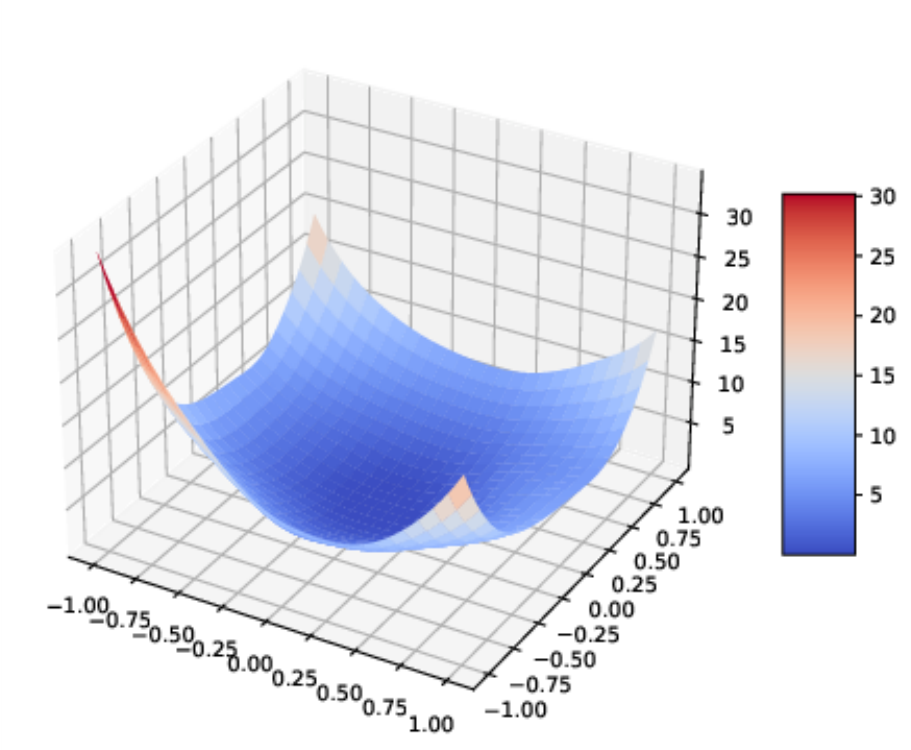}
    \caption{FP}
    \label{fig:cifar10_vgg8_fp_400epochs_train_loss_3dsurface}
  \end{subfigure}
  \begin{subfigure}{.32\linewidth}
    \centering
    \includegraphics[width=4.6cm]{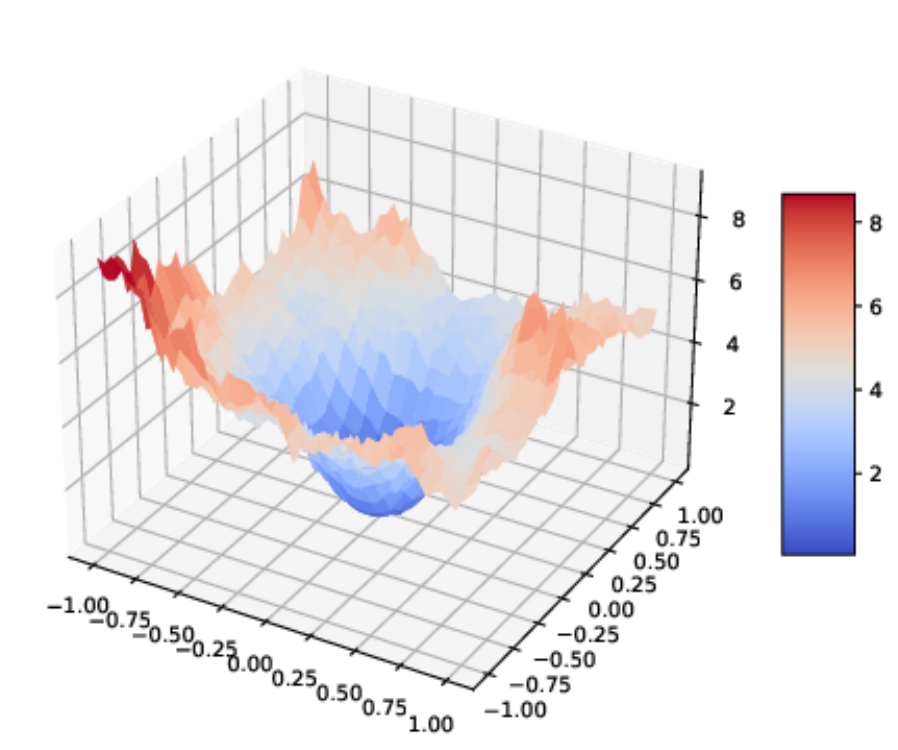}
    \caption{SQAKD (EWGS)}
    \label{fig:cifar10_vgg8_sqakd_400epochs_train_loss_3dsurface}
  \end{subfigure}
  \begin{subfigure}{.32\linewidth}
    \centering
    \includegraphics[width=4.6cm]{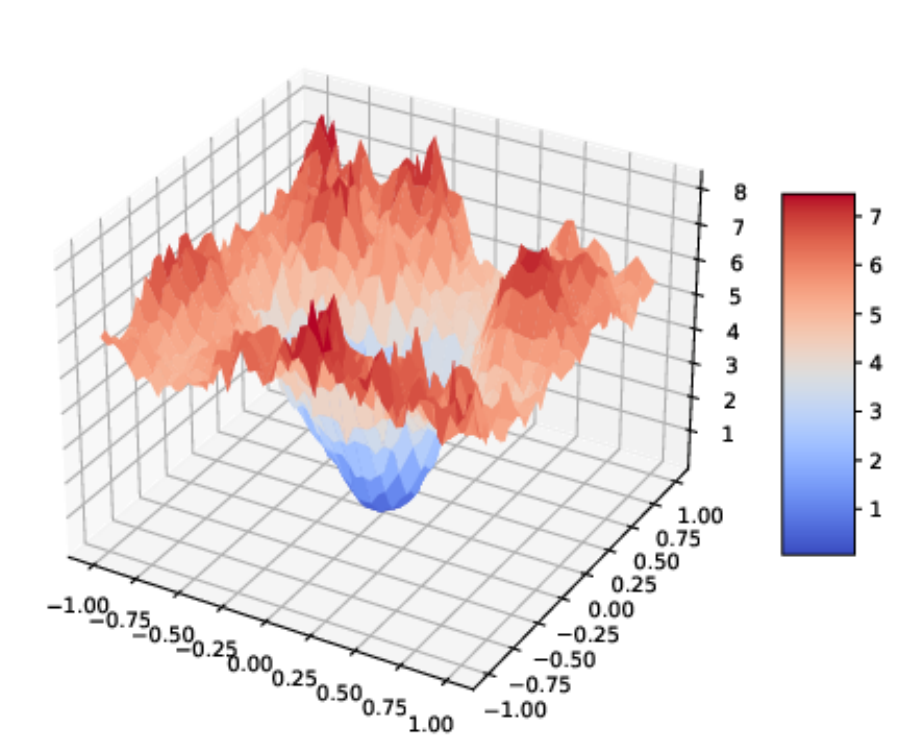}
    \caption{EWGS}
    \label{fig:cifar10_vgg8_ewgs_400epochs_train_loss_3dsurface}
  \end{subfigure}
    \begin{subfigure}{.32\linewidth}
    \centering
    \includegraphics[width=4.6cm]{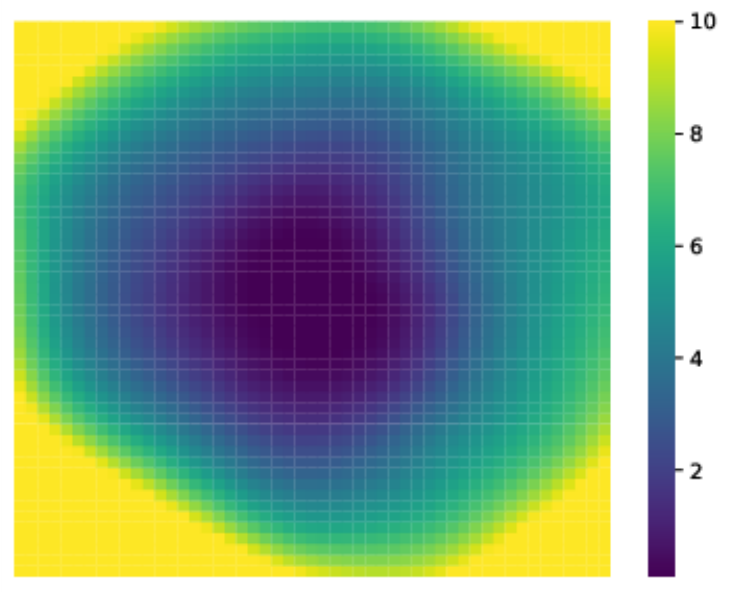}
    \caption{FP}
    \label{fig:cifar10_vgg8_fp_400epochs_train_loss_2dheat}
  \end{subfigure}
  \begin{subfigure}{.32\linewidth}
    \centering
    \includegraphics[width=4.6cm]{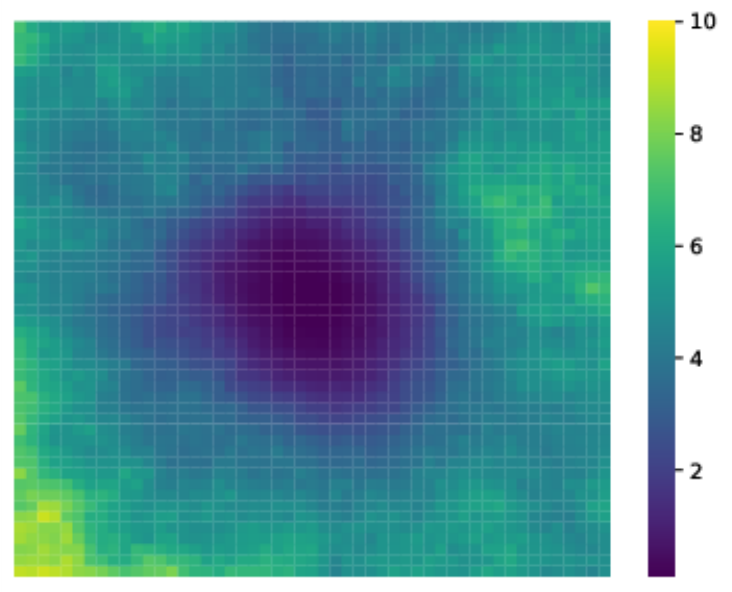}
    \caption{SQAKD (EWGS)}
    \label{fig:cifar10_vgg8_sqakd_400epochs_train_loss_2dheat}
  \end{subfigure}
  \begin{subfigure}{.32\linewidth}
    \centering
    \includegraphics[width=4.6cm]{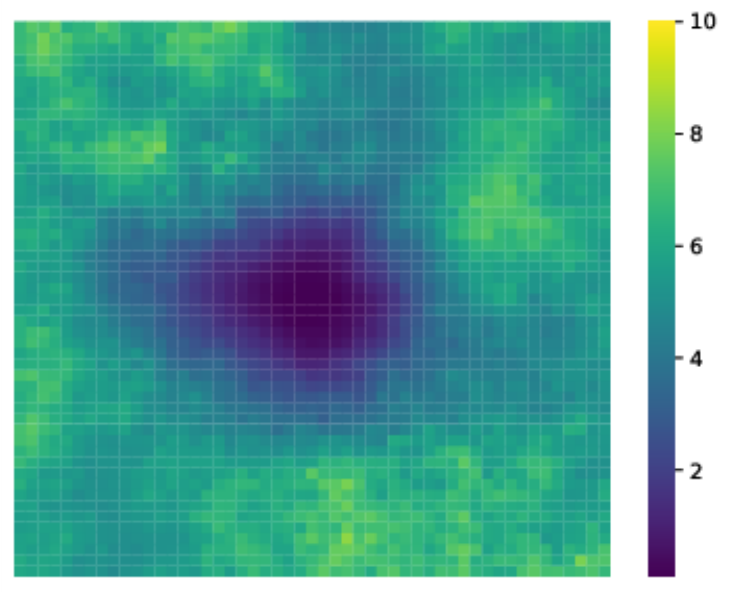}
    \caption{EWGS}
    \label{fig:cifar10_vgg8_ewgs_400epochs_train_loss_2dheat}
  \end{subfigure}
  %
    \begin{subfigure}{.32\linewidth}
    \centering
    \includegraphics[width=4.6cm]{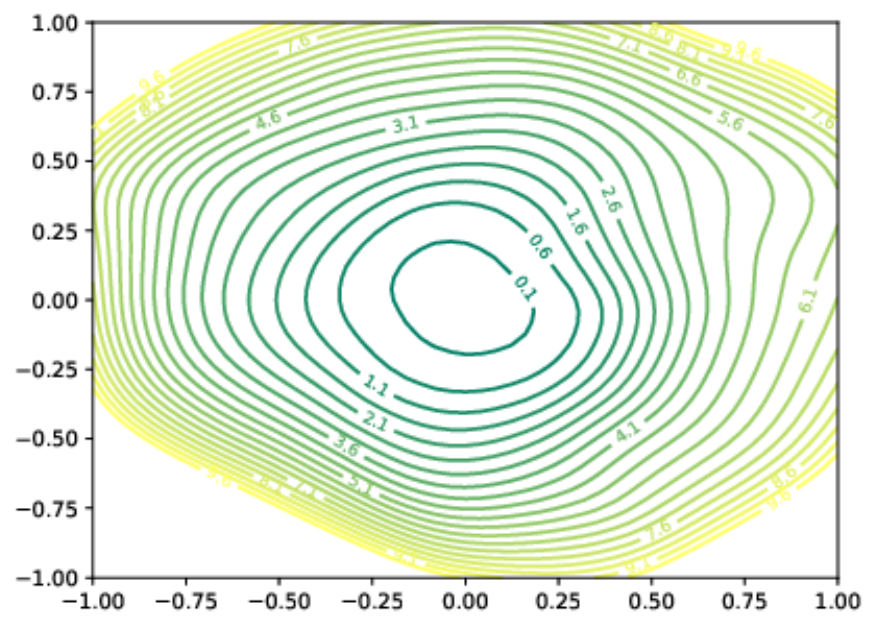}
    \caption{FP}
    \label{fig:cifar10_vgg8_fp_400epochs_train_loss_2dcontour}
  \end{subfigure}
  \begin{subfigure}{.32\linewidth}
    \centering
    \includegraphics[width=4.6cm]{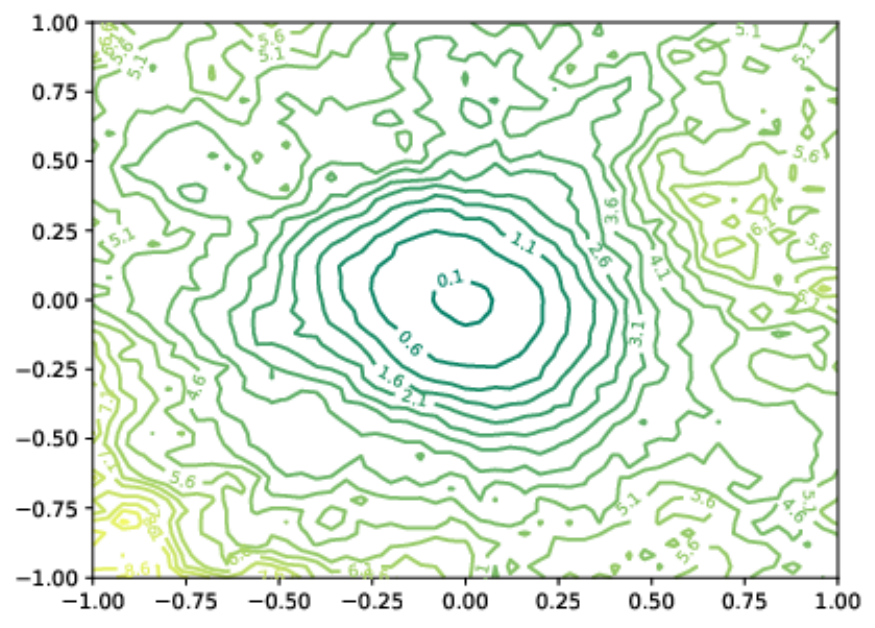}
    \caption{SQAKD (EWGS)}
    \label{fig:cifar10_vgg8_sqakd_400epochs_train_loss_2dcontour}
  \end{subfigure}
  \begin{subfigure}{.32\linewidth}
    \centering
    \includegraphics[width=4.6cm]{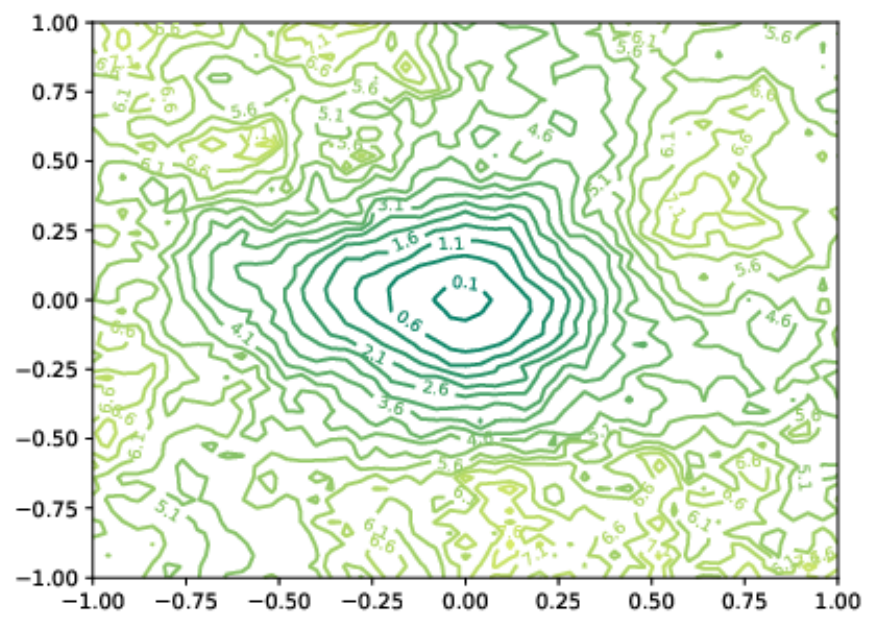}
    \caption{EWGS}
    \label{fig:cifar10_vgg8_ewgs_400epochs_train_loss_2dcontour}
  \end{subfigure}
    \begin{subfigure}{.32\linewidth}
    \centering
    \includegraphics[width=4.6cm]{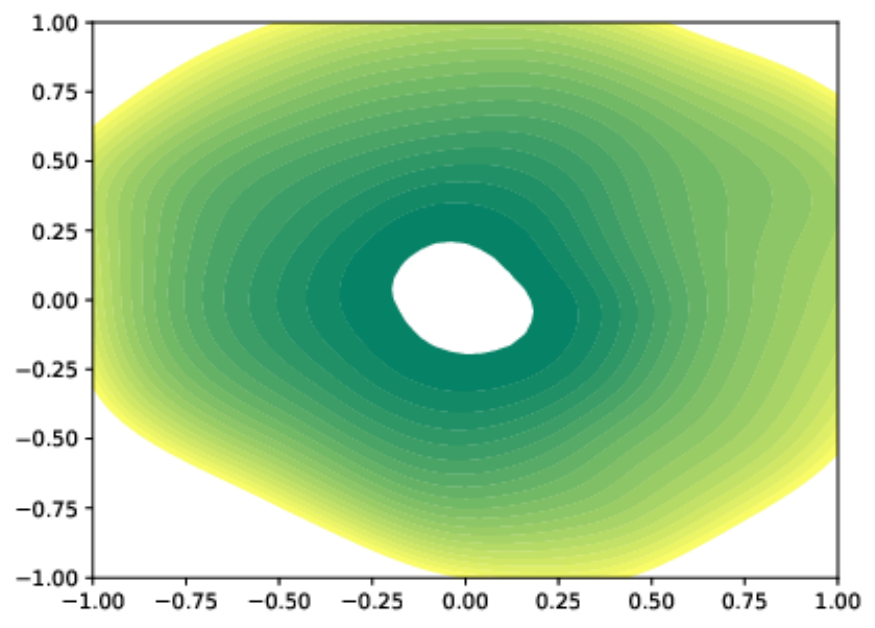}
    \caption{FP}
    \label{fig:cifar10_vgg8_fp_400epochs_train_loss_2dcontourf}
  \end{subfigure}
  \begin{subfigure}{.32\linewidth}
    \centering
    \includegraphics[width=4.6cm]{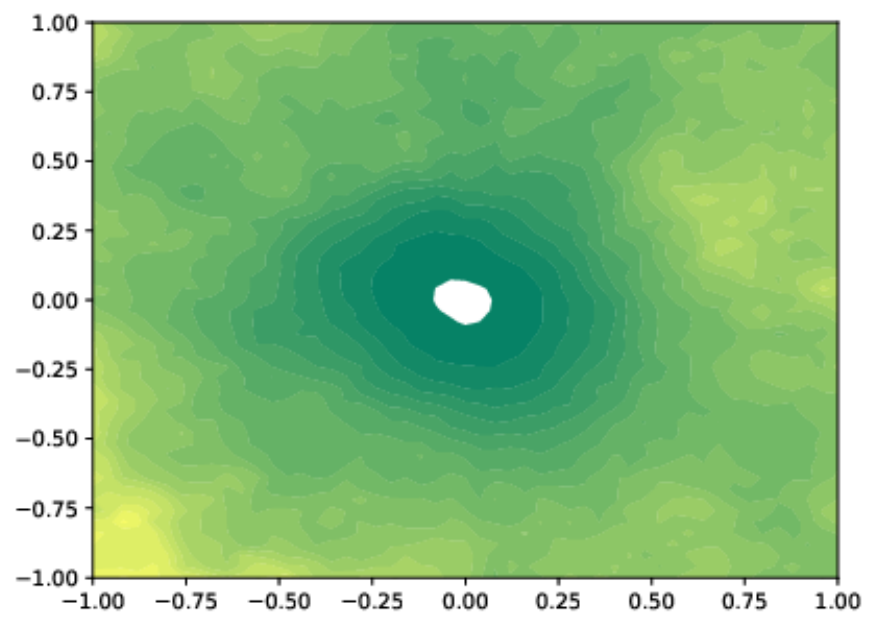}
    \caption{SQAKD (EWGS)}
    \label{fig:cifar10_vgg8_sqakd_400epochs_train_loss_2dcontourf}
  \end{subfigure}
  \begin{subfigure}{.32\linewidth}
    \centering
    \includegraphics[width=4.6cm]{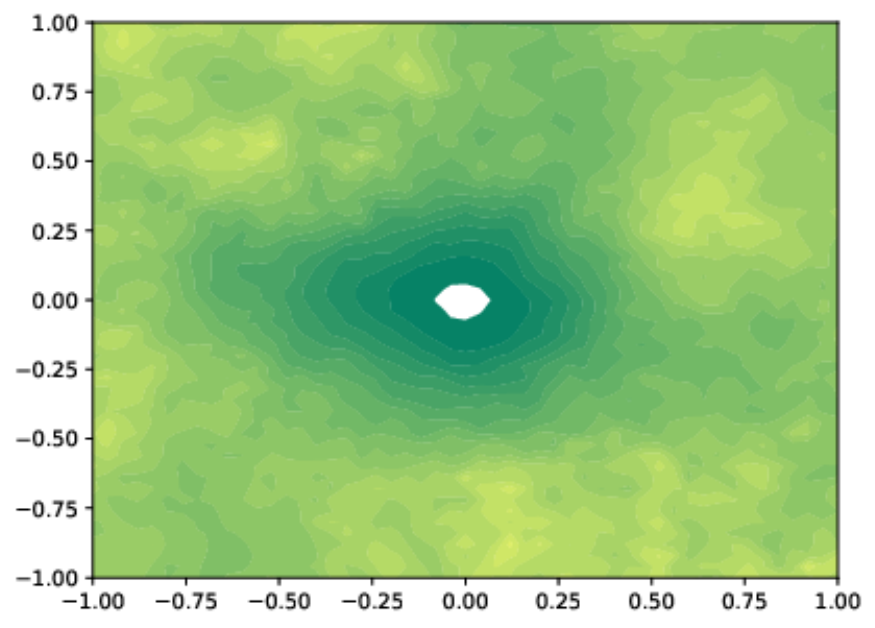}
    \caption{EWGS}
    \label{fig:cifar10_vgg8_ewgs_400epochs_train_loss_2dcontourf}
  \end{subfigure}
    \caption{3D loss surface (a, b, c), and its corresponding 2D heat representations (d, e, f), 2D contour visualizations (g, h, i), and 2D filled contour visualizations (j, k, l) for full-precision (FP), the standalone EWSG, and SQAKD integrating EWSG, on 2-bit VGG-8 with CIFAR-10.}
  \label{fig:ablation_loss_surface}
\end{figure}

Figure~\ref{fig:ablation_loss_surface} shows the 3D loss surface and the corresponding 2D heat, 2D contour, and 2D filled contour visualizations of the full-precision VGG-8 and the 2-bit VGG-8 trained using SQAKD and the standalone EWGS on CIFAR-10. 
SQAKD enables the quantized model to achieve a flatter and smoother loss surface compared to using the standalone EWGS.


\vfill


